\documentclass[10pt,twocolumn,letterpaper]{article}
\pdfoutput=1
\usepackage{cvpr}
\usepackage{times}
\usepackage{epsfig}
\usepackage{graphicx}
\usepackage{amsmath}
\usepackage{amssymb}

\usepackage{amsthm}
\usepackage[utf8]{inputenc}
\usepackage[english]{babel}
\usepackage{algorithm}
\usepackage{adjustbox}
\usepackage{import}
\usepackage[noend]{algpseudocode}
\usepackage{booktabs} 
\usepackage{relsize}
\usepackage{subcaption}
\usepackage{bm}
\usepackage{mathtools}
\usepackage{multirow}
\usepackage{array}

\newcommand{\grad}[2]{g_{#2}^{(#1)}}
\newcommand{\gradx}[2]{g_{#2}}

\newtheorem{lemma}{Lemma}
\newtheorem{innercustomthm}{Lemma}
\newenvironment{customthm}[1]
  {\renewcommand\theinnercustomthm{#1}\innercustomthm}
  {\endinnercustomthm}

\usepackage[pagebackref=true,breaklinks=true,letterpaper=true,colorlinks,bookmarks=false]{hyperref}

\cvprfinalcopy 

\def\cvprPaperID{7741} 

\ifcvprfinal\fi
\begin{document}
\title{\vspace{-0.5em}\text{\it iTAML} : An Incremental Task-Agnostic Meta-learning Approach}



 \author{
		Jathushan Rajasegaran$^{*}$, Salman Khan$^{*}$,  Munawar Hayat$^{*}$,  Fahad Shahbaz Khan$^{*}$,  Mubarak Shah$^{\dagger}$\\
\large $^{*}$Inception Institute of Artificial Intelligence, UAE,  \large $^{\dagger}$University of Central Florida, USA \tabularnewline
\texttt{\small \{first.lastname\}@inceptioniai.org}, \texttt{\small mshah@ucf.edu}  \tabularnewline
}

\maketitle
\thispagestyle{empty}

\begin{abstract}
\vspace{-0.3cm}
  Humans can continuously learn new knowledge as their experience grows. In contrast, previous learning in deep neural networks can quickly fade out when they are trained on a new task. In this paper, we hypothesize this problem can be avoided by learning a set of generalized parameters, that are neither specific to old nor new tasks. In this pursuit, we introduce a novel meta-learning approach that seeks to maintain an equilibrium between all the encountered tasks. This is ensured by a new meta-update rule which avoids catastrophic forgetting. In comparison to previous meta-learning techniques, our approach is task-agnostic. When presented with a continuum of data, our model automatically identifies the task and quickly adapts to it with just a single update. We perform extensive experiments on five datasets in a class-incremental setting, leading to significant improvements over the state of the art {methods} (e.g., a 21.3\% boost on CIFAR100 with 10 incremental tasks). Specifically, on large-scale datasets that generally prove difficult cases for incremental learning, our approach delivers absolute gains as high as  19.1\% and 7.4\% on ImageNet and MS-Celeb datasets, respectively. Our codes are available at: \small{\url{https://github.com/brjathu/iTAML}}\,.
\end{abstract}

\vspace{-0.5cm}
\section{Introduction}
\vspace{-0.1cm}
Visual content is ever-evolving and its volume is rapidly increasing each day. High-dimensionality and mass volume of visual media makes it impractical to store ephemeral or streaming data and process it all at once. Incremental Learning (IL) addresses this issue, and requires an agent to continually learn new tasks, while preserving old knowledge with limited or no access to previous data. The goal is to end-up with a single model, that performs well for all the tasks. In this manner, incremental learning models offer adaptation, gradual development and scalability capabilities that are very central to human-inspired learning.
\begin{figure}[!t]
    \centering
    \includegraphics[scale=1.3]{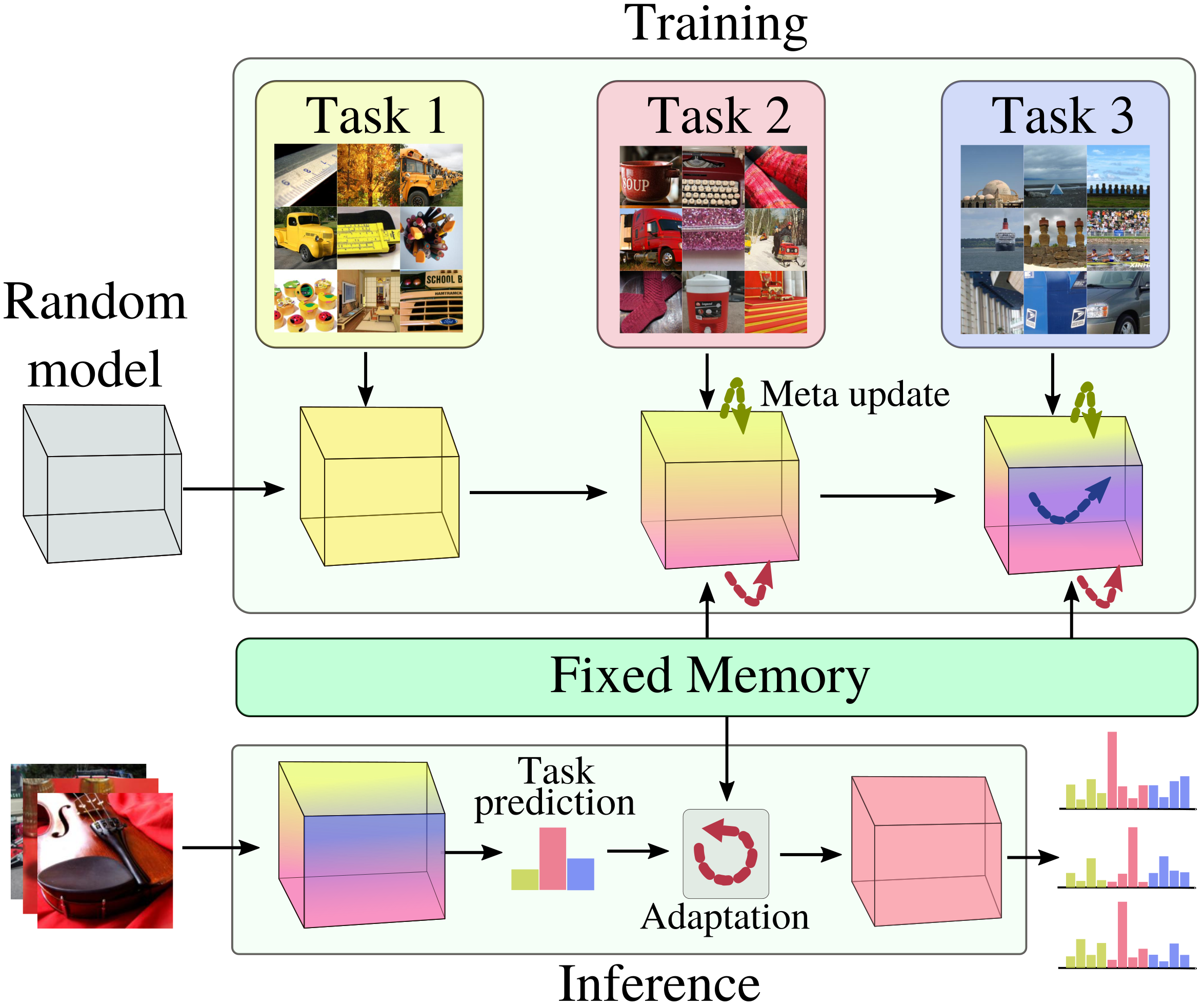}
    \vspace{-0.2cm}
    \caption{We propose a task-agnostic meta-learning approach for class-incremental learning. In this setting, tasks are observed sequentially where each task is a set of classes. During training, \emph{iTAML} incrementally learns new tasks with meta-updates and tries to retain previous knowledge to learn a generic model. At inference, given a data continuum, \emph{iTAML} first predicts the task and then quickly adapts to it. A fixed-size memory buffer is kept to support the adaptation.
    }
    \label{fig:front}
\end{figure}

This paper studies class-incremental learning where groups of classes are sequentially observed. This case is fundamentally different from conventional classification task due to two key factors, \emph{plasticity} and \emph{stability}. A \emph{plastic} network can quickly forget old tasks while performing well on new ones (forgetting), and a \emph{stable} network can impede learning new tasks in an effort to keep old information (interference). Despite several attempts, catastrophic forgetting and interference are still challenging problems for deep learning \cite{pfulb2019comprehensive}. This paper takes a different approach to IL based on the following principle: instead of a `one size fits all’ approach that learns a single model well-suited for all tasks, we propose to learn a generic meta-model which can be quickly adapted to the desired task. The meta-learning framework focuses on learning `how to rapidly learn?' The generality of the model arises from the `\emph{learning to learn}’ training strategy, that focuses on figuring out the shared parametric space common to all hitherto observed tasks. 

Adapting meta-learning for IL involves numerous challenges. \emph{First}, an incremental classifier keeps observing new tasks, therefore it must maintain a balance between plasticity and stability. Current meta-learning algorithms such as MAML (Model Agnostic Meta-Learning) \cite{finn2017model}, FOMAML (First Order MAML) \cite{nichol2018first} and Reptile~\cite{nichol2018reptile} do not offer the ability to adapt to new tasks or remember old tasks. \emph{Secondly}, without any task-specific feedback, meta-learning algorithms are not suitable for classification problems. Although meta-learned models can quickly adapt to new tasks, they require task-information to update their parameters which restricts their applicability. \emph{Finally}, most IL methods use a fixed-size memory buffer for old task replay. This makes the training distribution imbalanced due to majority samples from newer task, thereby learning a biased model. 

To overcome these challenges, we propose \textit{iTAML}, a task-agnostic meta-learning algorithm specifically designed for IL settings. To minimize catastrophic forgetting, \emph{iTAML} applies a meta-update rule that maintains an equilibrium between current and old knowledge. Further, it separates the generic feature extraction module from the task-specific classifier, similar to~\cite{javed2019learning}, thereby minimizing interference and promoting a shared feature space amongst tasks. To avoid the limitation of knowing task information before-hand, \textit{iTAML} learns a task-agnostic model that predicts the task automatically and subsequently adapts to the predicted task. This makes \textit{iTAML} unique compared to other meta-learning strategies. Finally, \textit{iTAML} mitigates the data imbalance problem by tuning the task specific parameters separately for each task. Thus, the task-specific parameters are not influenced by the majority samples of the current task.

The major contributions of this work are:
\textbf{(i)} A new task-agnostic meta-learning algorithm, \emph{iTAML}, which automatically predicts the task as well as the final class,
\textbf{(ii)}  A momentum based strategy for meta-update which effectively avoids forgetting,
\textbf{(iii)}  A new sampling rate selection approach that provides the lower-bound for the number of samples in the data continuum required for meta-update during inference, and
\textbf{(iv)} Significant performance gains demonstrated by extensive experiments on ImageNet, CIFAR100, MNIST, SVHN and {MS-Celeb} datasets.

\vspace{-0.1cm}
\section{Related Work}\vspace{-0.1cm}

Existing IL methods propose architectural modifications for deep CNNs e.g., dynamic networks \cite{rps_net,rajasegaran2019adaptive}, dual-memory modules \cite{gepperth2016bio}, and network expansion \cite{rusu2016progressive}. Rehearsal based methods have also been proposed that replay the old task by using an exemplar set \cite{rebuffi2017icarl,chaudhry2019efficient} or synthesize samples using generative models \cite{shin2017continual,sutton1990integrated}. IL approaches that work fundamentally on \textit{algorithmic level} 
can be grouped into regularization 
and  meta-learning based methods. We discuss these two sets of approaches next.

\noindent \emph{\textbf{Regularization Strategies for Incremental Learning:}}
The regularization based methods impose constraints during learning that seek to retain past knowledge. For example, learning without forgetting \cite{li2018learning} adds a distillation loss to preserve the old knowledge while sequentially learning new tasks. Different from the `task-incremental' setting explored in \cite{li2018learning}, \cite{rebuffi2017icarl,Castro_2018_ECCV} apply distillation loss in `class-incremental' setting to reduce forgetting. A distillation loss on the attention maps of the deep network is proposed in \cite{dhar2019learning} that minimizes overriding the old-task information. Recently, \cite{Wu_2019_CVPR} advocates for a simple bias correction strategy that promotes re-balancing the final classifier layer to give equal importance to current and older classes.

A broad set of regularization approaches introduce synaptic intelligence \cite{zenke2017continual, kirkpatrick2017overcoming,aljundi2018memory} which estimate the importance of each neuron and selectively overwrite the less-important weights as the learning progresses. Lee \etal \cite{lee2017overcoming} propose to incrementally learn new tasks by merging the old and new parameters via first and second order moment matching of posterior distributions. Elastic weight consolidation (EWC)~\cite{kirkpatrick2017overcoming} computes synaptic importance offline with a Fisher information matrix. It is used to slow down the learning for weights highly relevant to previous tasks. However, EWC can also cause intransigence towards new tasks, for which \cite{chaudhry2018riemannian} proposes exemplar rehearsal alongwith a Riemannian manifold distance measure for regularization.

\noindent \emph{\textbf{Meta-learning for Incremental Learning:}}
The overarching goal of meta-learning is to learn a model on a series of tasks, such that a new task can be quickly learned with minimal supervision. Meta-learning is thus ideally suited for IL since tasks are progressively introduced. A number of IL methods inspired by meta-learning have recently been proposed. Riemer \etal \cite{riemer2018learning} learn the network updates that are well-aligned and avoid moving in directions that can cause interference. 
However, \cite{riemer2018learning} uses a fixed loss function to align the gradients, that cannot be used in customized applications. Javed \etal \cite{javed2019learning} propose a meta-learning approach that disentangles generic representations from task-specific learning. Specifically, as new tasks are introduced, the task head is learned with both the inner and outer (meta) updates, while the representation learning backbone is only adapted with the outer (more-generic) updates. However, \cite{javed2019learning} assumes the training data is a correlated data stream, and all task samples are concurrently present during inference. Unlike \cite{javed2019learning}, we do not impose such strict constraints.  
Also, both of these methods~\cite{riemer2018learning,javed2019learning} update the inner loop by using a single sample at one time, which is not suitable for large-scale IL. Moreover, \cite{lopez2017gradient} assumes that the task is known for the data continuum which limits its applicability to practical scenarios. Jamal \etal~\cite{Jamal_2019_CVPR} present a task-agnostic meta-learning approach applicable only to few-shot learning. In contrast, \textit{iTAML} is task-agnostic and well suited for large-scale  settings. Our proposed meta-update is unbiased towards majority class samples and simultaneously minimizes forgetting. At inference, our model automatically adapts to the predicted task and uses task specific weights for class estimates. Besides, for the first time, we show the promise of meta-learning for large-scale incremental object recognition on five popular datasets.

\section{Proposed Method}

Our proposed \emph{iTAML} is a class IL approach that is model \& task-agnostic (i.e. independent of the network architecture and does not require task information). During training, we find a shared set of parameters that can work well for new tasks with minor local changes. \emph{iTAML} therefore learns generic meaningful representations that are transferable across tasks. In other words, meta-learning process forces the model to understand the inherent relationship between sequential tasks. At inference, given a data continuum with all samples belonging to the same task, our approach follows a two stage prediction mechanism. First, we predict the task using our generic model, then, we quickly adapt to the predicted task and find the class labels.

\subsection{Incremental Task Agnostic Meta-learning}

We progressively learn a total of $T$ tasks, with $U$ number of classes per task. Consider a classification model divided into two sub-nets, a feature mapping network $f_{\theta}$ and a classifier $f_{\phi}$. The function of both networks is given by, 
\vspace{-0.2cm}
\begin{align*}
    f_{\theta}: & \, \bm{x}\in\mathbb{R}^{C\times H\times W} \mapsto \bm{v}\in \mathbb{R}^{1 \times D} \\
    f_{\phi}: & \, \bm{v}\in \mathbb{R}^{1 \times D} \mapsto \bm{p}\in \mathbb{R}^{1 \times (UT)},
\end{align*}

\vspace{-0.2cm}
\noindent where, $f_{\theta}$ maps an input image  $\bm{x}$ to a feature vector $\bm{v}$, and $f_{\phi}$ maps $\bm{v}$ to output predictions $\bm{p}$. We start with a set of randomly initialized parameters $\Phi = \{\theta, \phi\}$, where $\phi = \left [\phi_1^\top, \dots , \phi_T^\top \right ]^\top$ and $\phi_i \in \mathbb{R}^{U \times D}$ are the task-specific classification weights. Training the first task is straightforward, however, when we get a new task $t \in [1,T]$, the old parameters $\Phi^{t-1}$ should generalize to all $t$ tasks. 

Our proposed meta-learning approach (Algorithm~\ref{alg:iTAML}) involves two updates, an \emph{`inner loop'}  which generates task-specific models for each task, and an \emph{`outer loop'} which combines task-specific models into a final generic model. \\
\textbf{Inner loop:} To train the inner loop, we randomly sample a mini-batch with $K$ triplets $\mathcal{B}_{m} = \{(\bm{x}_k, {y}_k, \ell_k)\}_{k=1}^{K}$ from the union set of current task training data $\mathcal{D}(t)$ and the exemplar memory $\mathcal{M}(t-1)$ containing a small number of samples for old tasks. Here, $\bm{x}_k,y_k$ and $\ell_k$ are the training images, class labels and task labels, respectively. This randomly sampled mini-batch contains training samples from multiple tasks. To train the task-specific model, we first group the training samples according to the tasks to form {a micro-batch} $\mathcal{B}^i_{\mu} = \{(\bm{x}_j,y_j, \ell_j)'\}_{j=1}^{J}$ per each task $i \in [1,t]$, where all $\ell_j$ in a micro-batch are identical. Inner loop parameters $\Phi_i =\{\theta, \phi_i\}$ are updated such that Binary Cross-entropy (BCE) loss is minimized on each micro-batch.
Here, $\theta$ is updated in the inner loop for all tasks, but $\phi_i$ is only updated for $i^{th}$ task. Also for each task, $\phi_i$ are updated for $r$ iterations using the same micro-batch. This helps in obtaining task-specific models closer to their original task-manifold, thereby providing a better estimate for gradient update in the outer-loop to obtain a generic model. \\
\begin{figure}[!tp]
    \centering
    \includegraphics[scale=0.6, angle=-90]{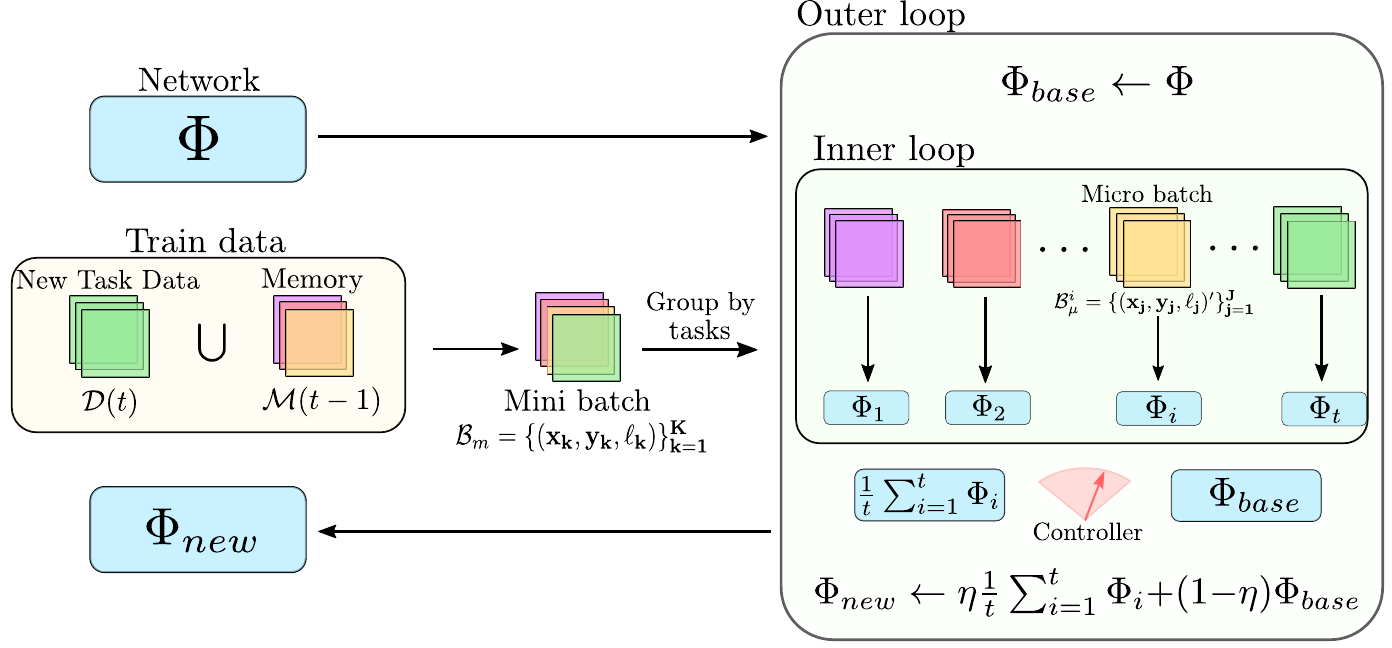}
    \vspace{-0.3cm}
    \caption{\emph{iTAML Meta-update}: Mini-batches are randomly sampled from a union set of new task training data and exemplar memory. Then, we group the samples according to the task, and create micro-batches which are used to generate task-specific models. Finally, in the outer loop all the task-specific models are combined. }
\end{figure}
\begin{algorithm}[b!]
\caption{Meta-training in \emph{iTAML}}
\label{alg:iTAML}
\algrenewcommand\algorithmicprocedure{}
\begin{algorithmic}[1]
\Procedure{}{}
\algorithmicrequire{ $\Phi^{t-1}$, $\mathcal{D}(t), \mathcal{M}(t-1)$, $t$, $T$ and $U$}
\State $\Phi \gets \Phi^{t-1} $ 
\For {$e$ iterations} 
    \State $\Phi_{base} \gets \Phi $
    \State $\mathcal{B}_m \sim \{\mathcal{D}(t) \cup \mathcal{M}(t-1)\}$
    \For {$i \in [1,t]$}
        \State $\Phi_i \gets  \{\theta, \phi_i\}$
        \State $ \mathcal{B}^i_{\mu} \gets \texttt{filter}(\mathcal{B}_m, i)$
        \For {$r$ steps}
            \State $\{\hat{y_j}\}_{j=1}^{J} \gets \Phi_i(\{\bm{x}_{j}\}_{j=1}^{J})$ 
            \State $loss \gets \sum_j \texttt{BCE}({y_j}, \hat{y_j})$
            \State $\Phi_i \gets \texttt{Optimizer}(\Phi_i, loss)$
        \EndFor
    \EndFor
    \State $\eta \gets \exp(-\beta\cdot \frac{i}{t})$
    \State $\Phi \gets \eta\cdot\frac{1}{t}\Sigma_i \Phi_i + (1-\eta)\cdot\Phi_{base}$
\EndFor
\State \Return $\Phi^{t} \gets \Phi$
\EndProcedure
\end{algorithmic}
\end{algorithm}
\noindent\textbf{Outer loop:} In the outer loop of \emph{iTAML}, we combine the task specific models generated during the inner loop to form a more generic model. Let, $\Phi_{base}$ is the model parameter set before inner loop updates. Then, we treat the combined effect of 
 all $(\Phi_{base}-\Phi_i)$ as the gradient update for the outer loop~\cite{nichol2018reptile}. Simply put, we move the meta-model from $\Phi_{base}$ towards the average direction of all task-specific updates from $\Phi_{base}$ in the inner loop using a dynamic controller $\eta$, 
 \vspace{-0.2cm}
\begin{align*}
\Phi &= \Phi_{base} {-} \eta \frac{1}{t}\sum_{i=1}^t (\Phi_{base}{-}\Phi_i) =  \eta \frac{1}{t}\sum_{i=1}^t\Phi_i {+} (1{-}\eta)\Phi_{base}.
\end{align*}
As the training progresses, the model must learn new tasks while simultaneously preserving previous information. In an ideal case, the model should adapt quickly at the early stage of the learning, while during the later tasks, it must avoid any drastic changes since a generic set of features is already learned. To impose this, we use a simple {momentum-based} dynamic controller $\eta$, which speeds up the learning at the beginning and slows it down towards the end. This {momentum-based} controller is given by $\eta = \exp (-\beta\frac{t}{T})$,  where $\beta$ is the decay rate set using a validation set. As an example, in the last task, model parameters move $e^{-\beta}$ times slower than the first task in the outer loop.  Controller is similar to having an adaptive learning rate or an adaptive optimizer~\cite{nichol2018reptile}, however, our controller depends on the number of tasks seen previously. This allows us to keep the right balance between plasticity and stability. 

\subsection{\emph{iTAML} vs. Other Meta Algorithms}
\emph{iTAML} is close to Reptile~\cite{nichol2018reptile} meta-learning algorithm. However, \emph{iTAML} fundamentally differs from Reptile~\cite{nichol2018reptile} in two aspects. First, our meta-update rule is different from Reptile, and incorporates a balancing factor that stabilizes the contribution from old and new tasks. Further, Reptile requires multiple inner loop updates ($r>1$), whereas \emph{iTAML} works well for $r\geq1$. We elaborate these properties below.
\begin{lemma}
\label{lemma:1}
Given a set of feature space parameters $\theta$ and task classification parameters $\phi = \{\phi_1, \phi_2, \dots, \phi_T\}$, after $r$ inner loop updates, \emph{iTAML}'s meta-update gradient for task $i$ is given by,
$
    \mathlarger{g}_{itaml}(i) = \gradx{0}{i,0} + \dots + \gradx{0}{i,r-1}, 
$
where, $\gradx{0}{i,j}$ is the $j^{th}$ gradient update with respect to $\{\theta,\phi_i\}$ on a  single micro-batch. \text{(see proof in the supplementary material)}
\end{lemma}

Compared to Reptile algorithm, which favors multiple batches to update inner loop, \emph{iTAML} requires only one batch through all updates in the inner loop. As mentioned in \cite{nichol2018reptile}, $\mathlarger{g}_{reptile} = \grad{0}{i,0} + \grad{1}{i,1} + \dots + \grad{r-1}{i,r-1}$. Here, $\grad{m}{i,m}$ is the gradient calculated on $m^{th}$ disjoint micro-batch. This differs from our meta-update rule which relies on one micro-batch per task in the inner loop as compared to $r$ disjoint micro-batches per task in Reptile. We empirically found that a Reptile style meta-update is not useful for IL while our proposed {update} rule helps in finding task-specific weights useful for an optimal outer-loop update. This because, in an exemplar based IL setting, the memory limit per task decreases with new tasks. Hence, in a random sample of a mini-batch, old classes are under-represented compared to the new task. To do multiple micro-batch updates per task as in Reptile, we need to break a micro-batch further, which results in more noisy gradient updates. Therefore, \emph{iTAML} efficiently uses a single micro-batch per task.  
\begin{lemma}
\label{lemma:2}
Given a set of feature space parameters $\theta$ and task classification parameters $\phi = \{\phi_1, \phi_2, \dots \phi_T\}$, \emph{iTAML} allows to keep the number of inner loop updates $r\geq1$. \text{(see proof in the supplementary material)}
\end{lemma}
\noindent The above property shows that a single inner-loop update does not result in normal joint training for \emph{iTAML}. Thus, unlike Reptile, we can quickly meta-update with $r{=}1$. 

\begin{figure}[t!]
    \centering
    \includegraphics[trim= 1cm 0 0.2cm 0, clip, scale=0.23, angle=-90]{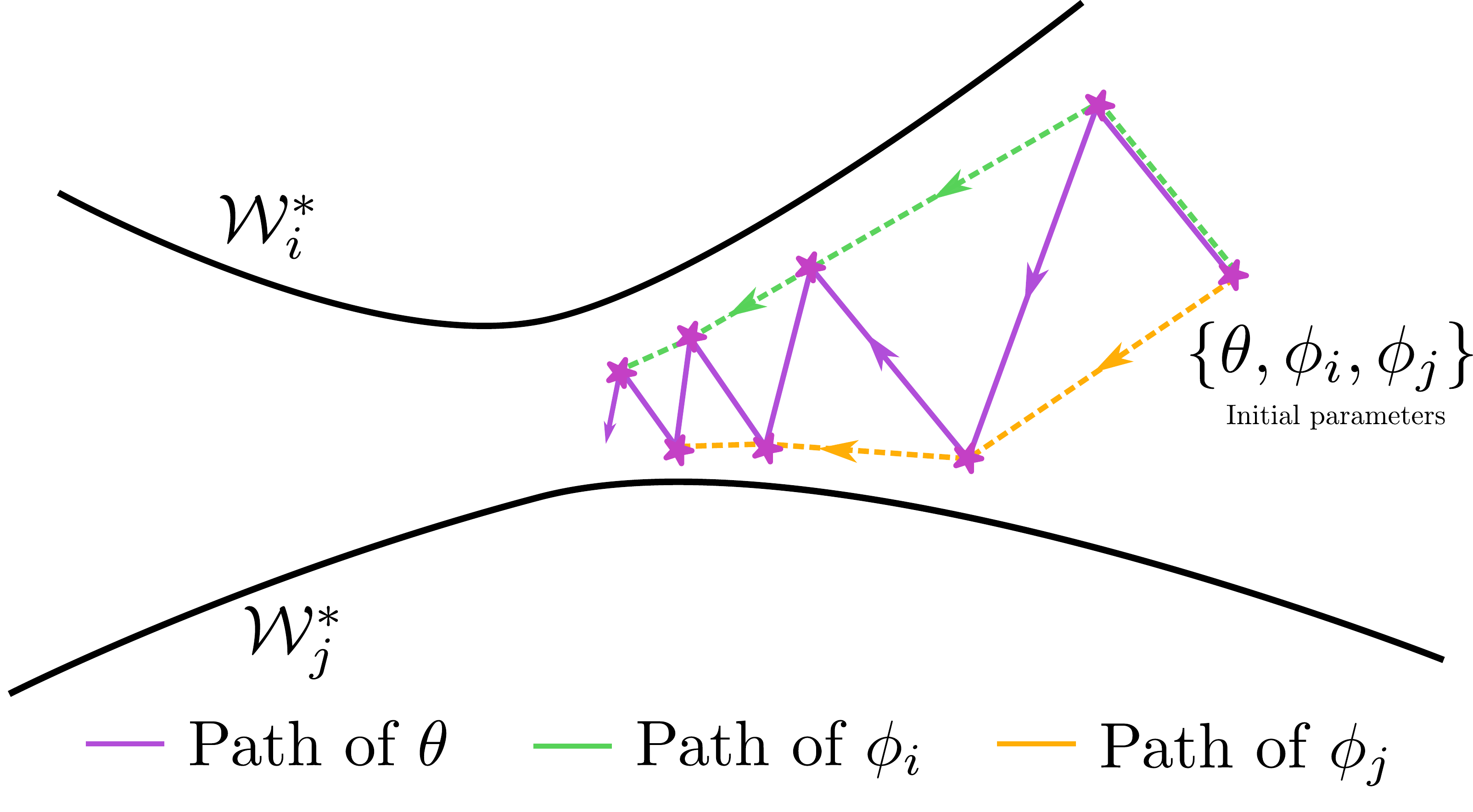}
    \vspace{-0.3cm}
    \caption{An illustration of how \emph{iTAML} gradient updates move the model parameters. Let $\mathcal{W}^*_i, \mathcal{W}^*_j$ be the optimal set of parameters 
    for tasks $i$ and $j$. \emph{iTAML} moves the feature space parameters $\theta$ towards the closet point between two optimal solution manifolds (solid line), while the classification parameters $\phi_i, \phi_j$ move only if $\theta$ moves towards its corresponding manifold (dashed lines). Therefore, the task specific classification parameters stay close to their optimal solution manifolds, which allows the model to predict tasks even without any gradient updates after meta-training.}
    \label{fig:manifold}
\end{figure}

Since \emph{iTAML} is task agnostic, even after outer loop update, it can predict tasks without requiring any external inputs. This is in contrast to existing meta-learning algorithms which can not be employed in supervised classification tasks, without requiring at least some fine-tuning. For example, in few-shot learning, meta-model parameters can classify a new task only after they are updated for the given support set. Therefore, for a generalized meta-model, such as Reptile~\cite{nichol2018reptile} and FOMAML~\cite{nichol2018first}, without any task information or support set, the meta-model parameters are less useful. This is because all the model parameters are updated in the inner loop for these methods. In comparison, for the case of joint training, model parameters are optimized using the current data available (i.e., exemplars and new task data) in a normal fashion. In terms of meta-learning, this is equivalent to a single gradient update using all task samples in a mini-batch. In contrast, for our proposed \emph{iTAML}, the classification parameters $\phi=\{\phi_0, \phi_1, \dots \phi_T\}$ are updated individually in the inner loop per task, and they remain task-specific even after the meta-update. This can be further explained from an optimal solution manifolds perspective (Fig.~\ref{fig:manifold}). Reptile and FOMAML move all the parameters towards a point on $\mathcal{W}^*$ which is close to all the task-specific optimal solution manifolds. In contrast, \emph{iTAML} only moves the feature space parameters $\theta$ towards $\mathcal{W}^*$ and keeps the classification parameters $\phi_t$ close to their corresponding optimal solution manifolds. Further, the fixed-sized exemplar memory results in an imbalanced data distribution. Due to this, Reptile, FOMAML and Joint training methods become more biased towards the later tasks. Since, \emph{iTAML} updates the classification parameters separately, it inherently overcomes the bias due to imbalance in tasks.

The above properties empower \emph{iTAML} to accurately predict class labels (close to joint training) without requiring any gradient updates at inference. Further, with a given data continuum, \emph{iTAML} can predict tasks with up to $100\%$ accuracy. This allows us to design a two-stage classifier, where we first predict the task of the data continuum without any additional external knowledge, and once the task is found, we apply a gradient update to convert the generalized weights to task-specific weights, using a small exemplar set. 

\subsection{\emph{iTAML} Inference }

\begin{figure}[!t]
    \centering
    \includegraphics[scale=0.73, angle=-90]{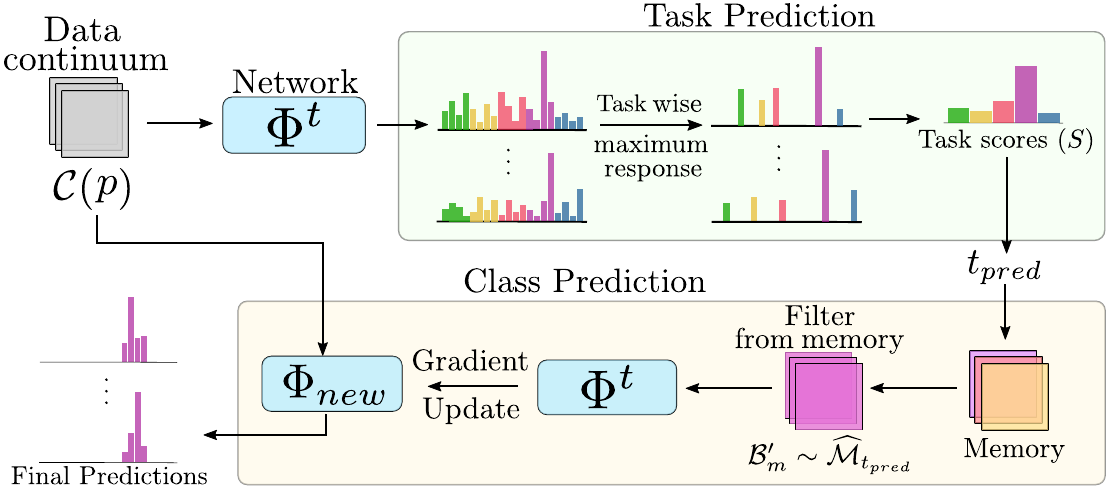}
    \vspace{-4pt}
    \caption{\emph{Task and Class prediction}: Given the data continuum $\mathcal{C}(p)$, for all samples the maximum responses per task are accumulated into a task score to get task prediction $t_{pred}$. For the class prediction, exemplars of task $t_{pred}$ from memory $\mathcal{M}$ are used to update the generic model $\Phi^t$ to task-specific model $\Phi_{new}$. The data continuum is then fed through the $\Phi_{new}$ to get the sub-class predictions. }
    \label{fig:class_pred}
\end{figure}

At inference time, we receive data as a continuum $\mathcal{C}(p) = \{\bm{x}_j: \ell_j = m\}_{j=1}^{p}$ for an unknown task $m$ with $p$ samples.  A data continuum is simply a group of samples of an identical task bundled together. Given $\mathcal{C}(p)$, inference happens in two stages. First, the task is predicted using generalized model parameters, and then these generalized parameters are updated to the task-specific parameters to predict classes belonging to the respective task. Fig.~\ref{fig:class_pred} outlines the flow of task and class prediction.

\noindent \textbf{Task Prediction:} Consider the model $\Phi^t$ trained for $t$ tasks, with $U$ classes in each task. Task prediction is straightforward. First, for each sample in the continuum, we get the final classification layer response. Then, for each response vector, a maximum response per task is recorded. An average of the maximum responses per task is used as the task score. A task with a maximum score is finally predicted $(t_{pred})$. Algorithm~\ref{alg:task_pred} outlines the task prediction steps.

\begin{algorithm}[h]
\caption{Task Prediction}
\label{alg:task_pred}
\algrenewcommand\algorithmicprocedure{}
\begin{algorithmic}[1]
\Procedure{}{}
\algorithmicrequire{ $\Phi^{t}$, $\mathcal{C}(p)= \{\bm{x}_j\}_{j=1}^{p}$ and $U, T$}
\State $S \gets [0, 0, 0, ... , 0]$  \Comment{initialize scores }
\For {$j = [1,2, \dots ,p]$} 
    \State $\hat{y}^j \gets \Phi^t(\bm{x}_j)$
    \For {$i = [1,2, \dots ,t]$} 
        \State $S[i] \gets S[i] + \max(\hat{y}^j[i\cdot U: (i+1) \cdot U])$ 
    \EndFor
\EndFor
\State \Return $t_{pred} \gets \text{argmax}(S)$
\EndProcedure
\end{algorithmic}
\end{algorithm}

\noindent \textbf{Class Prediction:} Class prediction involves updating the generalized parameters $\Phi^t$ using exemplars. To correctly predict the classes in a task, we move the generalized parameters towards task-specific parameters $\Phi_{new}$. To do so, we select samples from exemplar memory corresponding to classes of the predicted task $t_{pred}$. We use these labelled samples to do a single gradient update on the generalized parameters, which results in a task-specific parameters set.  The data continuum is then fed through these parameters to find the sub-classes of the predicted task. The final class is then derived as $(U \cdot t_{pred})   + \textit{subclass}$. Without losing generality, this can be extended to the case of uneven classes in different tasks. Algorithm~\ref{alg:class_pred} summarizes the main steps of the class prediction on data continuum. 
\begin{algorithm}
\caption{Class Prediction}
\algrenewcommand\algorithmicprocedure{}
\label{alg:class_pred}
\begin{algorithmic}[1]
\Procedure{}{}
\algorithmicrequire{ $\Phi^t$, $\mathcal{C}(p)$, $t_{pred}$ and memory $\mathcal{M}(t)$}
\State $\mathcal{\widehat{M}}_{t_{pred}} \gets \texttt{filter}(\mathcal{M}(t), t_{pred})$
\State $\Phi_{new} \gets \{\theta, \phi_{t_{pred}}\}$
\For {$b$ iterations}
    \State $\mathcal{B}'_m \sim \mathcal{\widehat{M}}_{t_{pred}}$ \Comment{mini-batch with $Q$ samples}
    \State $\{\hat{y}_q\}_{j=1}^{Q} \gets \Phi_{new}(\mathcal{B}'_m )$
    \State $loss \gets \sum_{j}\texttt{BCE}( y_j, \hat{y}_j)$
    \State $ \Phi_{new} \gets \texttt{Optimizer}(\Phi_{new}, loss)$
\EndFor
\For {$j \in  [1, p]$} 
    \State $\hat{y}_j \gets \Phi_{new}(\bm{x}_j)$
    \State \small{$\textit{subclass} \gets \texttt{argmax}(\hat{y}_j[U\cdot t_{pred}:U(t_{pred}+1)])$}
    \State  $C_{pred}^j \gets t_{pred}\cdot U + \textit{subclass}$
\EndFor
\State \Return $\bm{C}_{pred} = \{C_{pred}^j\}_{j=1}^{p}$
\EndProcedure
\end{algorithmic}
\end{algorithm}

\subsection{Limits on the Data Continuum Size}
At inference, the model is fed with a data continuum, which is used to identify the task. The model then adapts itself for the predicted task. The number of samples in the continuum plays a key role in task prediction. A higher number of samples attenuates the noisy predictions and results in a higher {task} accuracy. However, in practical settings, continuum sizes are limited. Therefore, it is necessary to know a lower bound on the continuum size to keep the task prediction accuracy at a certain (desired) level. 



Let the model's minimum prediction accuracy be $P_0$ after learning $t$ tasks, each with $U$ classes. $P_0$ can be interpreted as, $P_0 = \mathcal{P}(\mathcal{Z}=i|y_{true}=i)$, where event $\mathcal{Z}=i$ denotes the case when the maximum response occurs at class $i$, and $y_{true}$ is the true label. If $\bar{\mathcal{Z}}=i$ denotes the event when the maximum response occurs anywhere but at class $i$, then $\mathcal{P}(\bar{\mathcal{Z}}=i|y_{true}=i) = 1-P_0$.
Since \emph{iTAML} promotes a model to be unbiased, we assume that incorrect predictions are uniformly distributed. 
Thus, the probability that the maximum response happens at the correct task is,
\vspace{-0.2cm}
\begin{align*}
     \mathcal{P}(\mathcal{Z}=S_{\ell}|y_{true}=i) &= P_0 + \frac{1-P_0}{ U \cdot t-1} \cdot (U-1).
\end{align*}
Here, $S_{\ell}$ is the set containing the classes corresponding to the $\ell^{th}$ task (including class $i$), where $\ell = \texttt{floor}(\frac{i}{U})$. Hence, $\mathcal{P}(\mathcal{Z}=S_{\ell}|y_{true}=i)$ is the probability that the maximum response falls at any class of the corresponding task. Let $\hat{P}_0=\mathcal{P}(\mathcal{Z}=S_{\ell}|y_{true}=i)$ be the probability of correctly predicting the task. Now, if we have $n$ samples in a continuum then the probability of overall task prediction follows a binomial distribution. In $n$ samples correspond to any  one of the $t$ tasks, the random prediction would be $\texttt{round}(\frac{n}{t})$ samples belonging to any single task. Therefore, we require at least $\texttt{round}(\frac{n}{t})+1$ correctly predicted samples to find the task of a given continuum,
\vspace{-0.2cm}
\begin{align*}
\small
     \mathcal{P}(\text{correct task}) &= \sum_{k=\texttt{round}(\frac{n}{t})+1}^n {n\choose k} (\hat{P_0})^{k} (1-\hat{P_0})^{n-k}
\end{align*}
Our goal is to find a minimum value of $n$ such that, $\mathcal{P}(\text{correct task})> \gamma$, where, $\gamma$ is the required task accuracy level. Algorithm~\ref{alg:sample} explains the main steps of finding minimum value of $n$. Note that Algorithm~\ref{alg:sample} can easily be solved using any simple brute-force method.

\begin{algorithm}
\caption{Find Lower bound on Data Continuum Size}
\label{alg:sample}
\algrenewcommand\algorithmicprocedure{}
\begin{algorithmic}[1]
\Procedure{}{}
\algorithmicrequire{ Minimum class prediction accuracy $P_0$, required task accuracy $\gamma$, $U$ and $t$}
\State $N \gets t\cdot U$  \Comment{number of classes seen}
\State $\hat{P}_0  \gets \frac{U-1}{N-1}\cdot (1-P_0)$
\State $\min(n)$ subject to,
\State \ \ \ \ \ \ \ $\sum_{k=\texttt{round}(\frac{n}{t})+1}^n {n\choose k} (\hat{P_0})^{k} (1-\hat{P_0})^{n-k} > \gamma$
\State \Return $n$
\EndProcedure
\end{algorithmic}
\end{algorithm}



\definecolor{sh_gray}{rgb}{0.84,0.84,0.84}
\definecolor{sh_gray2}{rgb}{1,0.89,0.75}
\definecolor{color3}{rgb}{0.95,0.95,0.95}
\definecolor{color4}{rgb}{0.96,0.96,0.86}

\newcommand{\best}[1]{\colorbox{sh_gray2}{\textbf{#1}}}%
\newcommand{\sbest}[1]{\colorbox{sh_gray}{\textbf{#1}}}%
\vspace{-0.2cm}
\section{Experiments and Results}

\subsection{Implementation Details}

\noindent\textbf{Datasets:} We evaluate our method on a wide spectrum of incremental learning benchmarks. These include small scale datasets i.e., split MNIST~\cite{zenke2017continual} and split SVHN, where each task is assigned with two classes. For medium scale datasets, we use CIFAR100 and ImageNet-100, each divided into $10$ tasks (with $10$ classes per task). For large scale datasets, ImageNet-1K and MS-Celeb-10K with $1000$ and $10000$ classes respectively are used. MS-Celeb-10K is a subset of MS-Celeb-1M dataset~\cite{guo2017one}. We consider 10 tasks, each having $100$ and $1000$ classes per task respectively for ImageNet-1K and Celeb-10K. To be consistent with~\cite{Hsu18_EvalCL}, we keep a randomly sampled exemplar memory of $2000$ samples for MNIST, SVHN, CIFAR100 and ImageNet-100, $20$K for ImageNet-1K~\cite{rebuffi2017icarl} and $50$K for Celeb-10K~\cite{Wu_2019_CVPR}.

\noindent\textbf{Network Architectures:} For Split-MNIST, a simple two layer MLP (400 neurons each) similar to \cite{Hsu18_EvalCL} is used. For SVHN and CIFAR100, a reduced version of ResNet-18 (\emph{ResNet-18(1/3)}) is used, with the number of filters in all layers reduced by three times~\cite{lopez2017gradient}, resulting in $1.25\ million$ parameters. For ImageNet-100 and ImageNet-1K, we use standard ResNet-18. For training, we use RAdam~\cite{liu2019variance} with initial learning rate of $0.01$ for $70$ epochs. Learning rate is multiplied by $\frac{1}{5}$ after $20,40$ and $60$ epochs. All the models are trained on a single Tesla-V100 GPU.
\vspace{-0.1cm}
\subsection{Results and Comparisons}
\begin{figure*}[t!]
    \centering
    \begin{subfigure}{0.33\textwidth}
    \label{fig:meta_agnostic}
        \centering
        \includegraphics[width=0.99\textwidth]{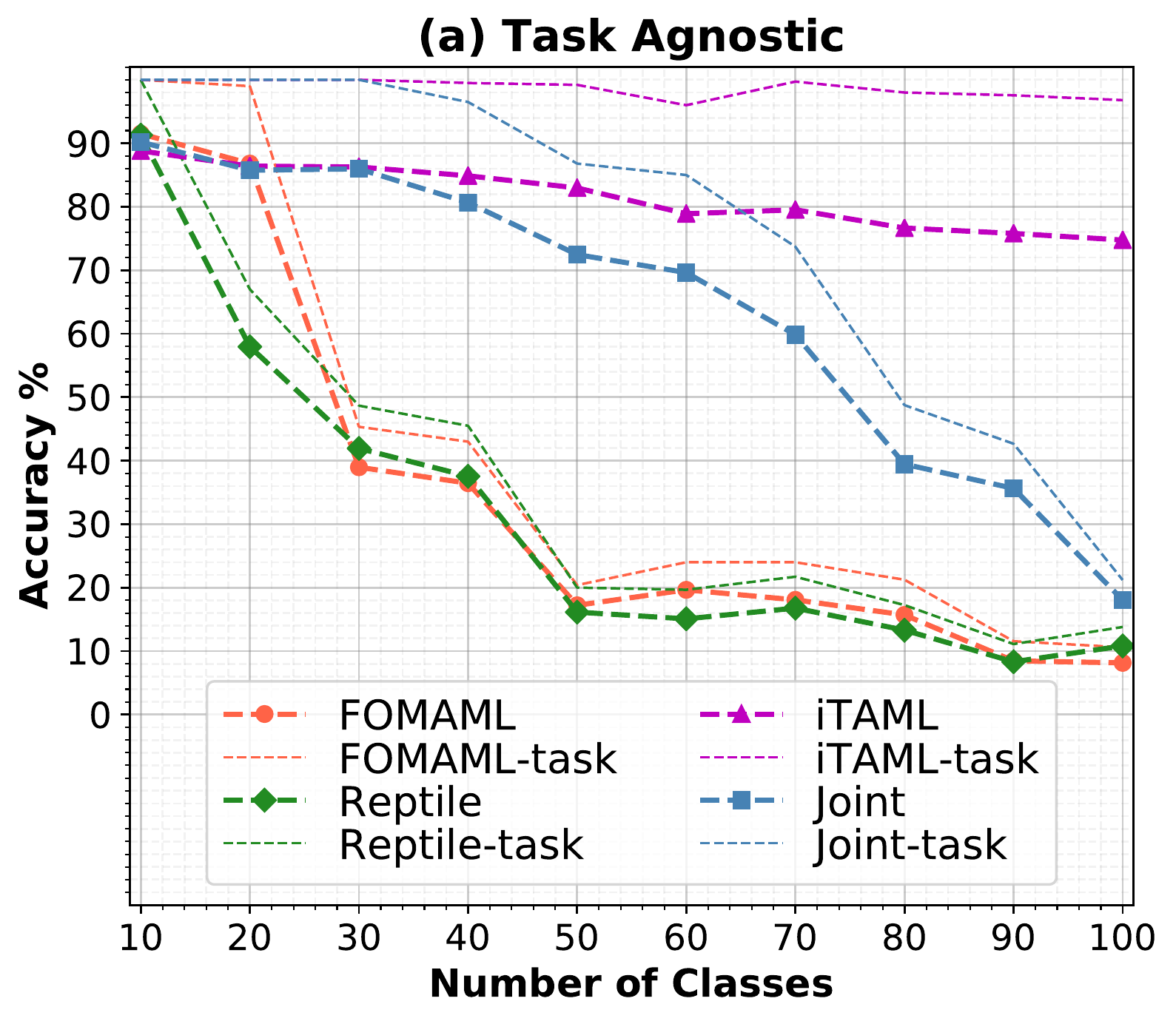}
        \vspace{-0.5cm}
    \end{subfigure}%
    \begin{subfigure}{.33\textwidth}
        \centering
        \includegraphics[width=0.99\textwidth]{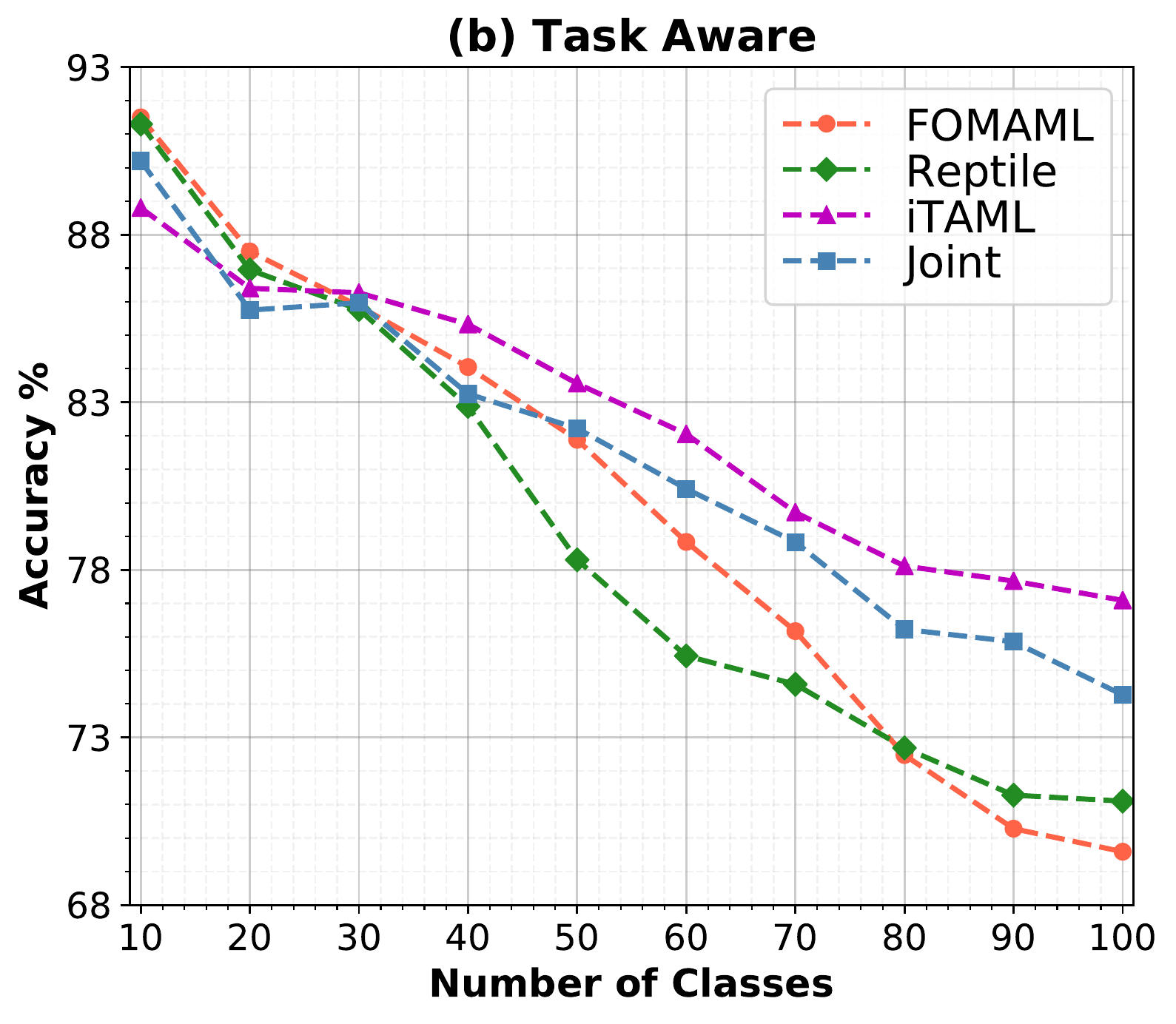}
        \vspace{-0.5cm}
        \label{fig:meta_aware}
    \end{subfigure}%
     \begin{subfigure}{0.33\textwidth}
        \centering
        \includegraphics[width=0.99\textwidth]{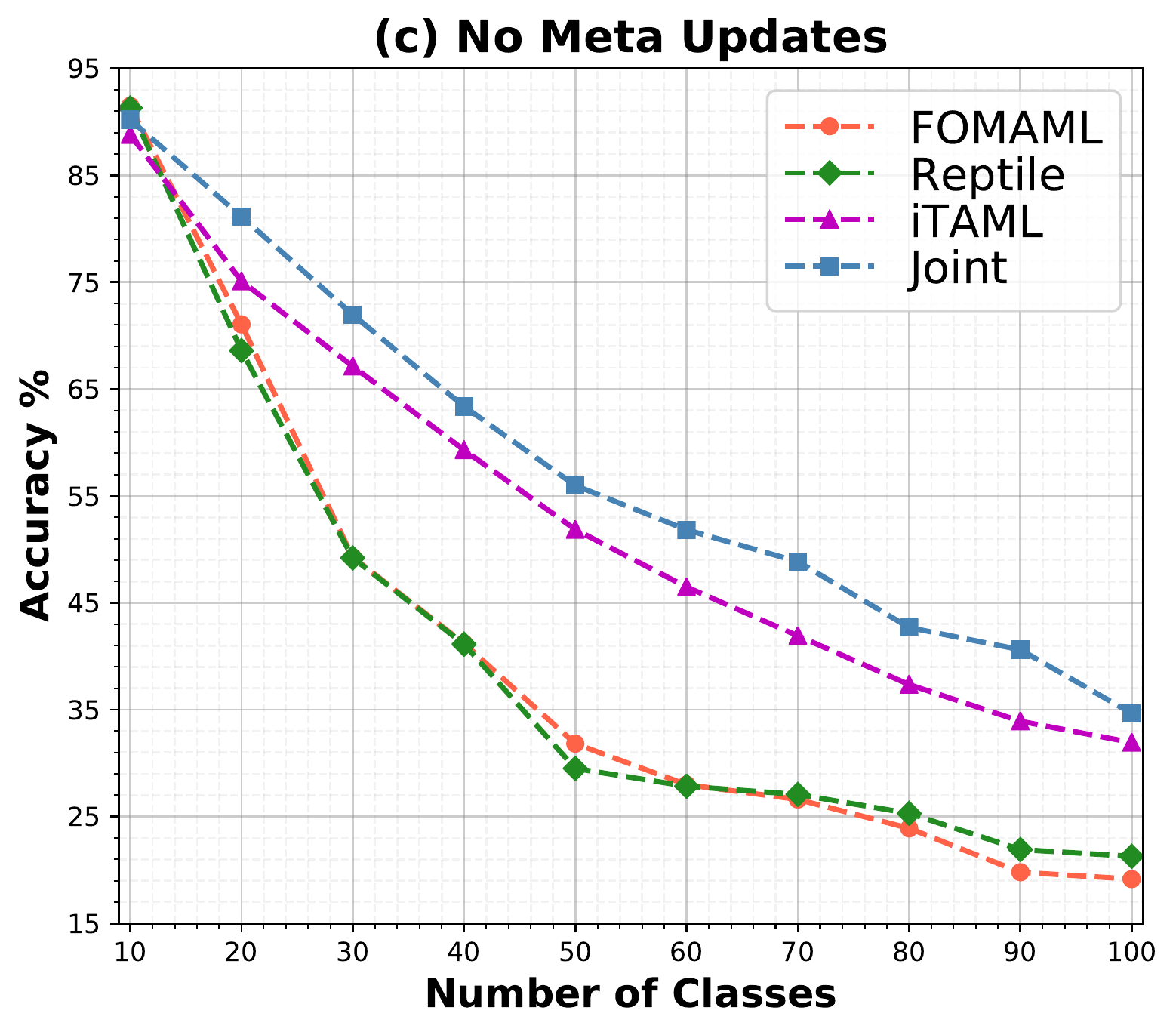}
        \vspace{-0.5cm}
        \label{fig:meta_claasification}
    \end{subfigure} 
    \vspace{-0.3cm}
    \setlength{\belowcaptionskip}{-12pt}
    \caption{\emph{Comparison between different meta-learning algorithms:} 
    {a)} task-agnostic,  {b)} task-aware, and  {c)} with no 
    inference updates. \emph{iTAML} performs the best not only in task-agnostic case, but also on task-aware  and no meta-update cases.}
    \label{fig:meta_algorithm}
\end{figure*}
\noindent \textbf{Comparison with meta-learning Algorithms:}~
Fig.~\ref{fig:meta_algorithm} compares different first-order meta-learning algorithms i.e., FOMAML and Reptile with our \emph{iTAML} and joint training on task-agnostic, task-aware, and no {inference-update} settings on CIFAR100 for $10$ tasks. In the task-agnostic setting, only the data continuum is present at inference, whereas for the task-aware setting, both data continuum and the task label are available. For the task-agnostic settings, the classification accuracy of Reptile and FOMAML drops drastically with more tasks since they are unable to precisely predict the tasks. Apart from that, joint training performance also drops after the first few tasks, mainly caused by highly imbalanced data distribution. However, \emph{iTAML} is able to consistently predict the tasks with above $95\%$ accuracy, while keeping an average classification accuracy of $77.79\%$ even after $10$ tasks. \emph{iTAML} inherently trains task-agnostic models, thus even after the meta-updates, it can predict the tasks independently. However, other first-order meta-learning algorithms require additional information (e.g. feedback from the task or the task label). Compared with task-agnostic settings, the performance of FOMAML and Reptile improves when the task label is known (task-aware settings). Nevertheless, for the task-aware settings, \emph{iTAML} still shows $6.9\%$ improvement over Reptile, since it effectively tackles stability-plasticity dilemma and imbalance problem. Fig.~\ref{fig:meta_algorithm} also compares different algorithms without performing any task-specific meta-updates. \emph{iTAML} performs similar to the joint training under these settings. 

\noindent\textbf{Comparison with existing methods: }We extensively compare \emph{iTAML} with several popular incremental learning methods, including Elastic Weight Consolidation~\cite{kirkpatrick2017overcoming}, Riemannian Walk (RWalk)~\cite{chaudhry2018riemannian}, Learning without Forgetting (LwF)~\cite{li2018learning}, Synaptic Intelligence (SI)~\cite{zenke2017continual}, Memory Aware Synapses (MAS)~\cite{aljundi2018memory}, Deep Model Consolidation (DMC)~\cite{zhang2019class}, Incremental Classifier and Representation Learning (iCARL)~\cite{rebuffi2017icarl}, Random Path Selection network (RPS-net)~\cite{rps_net} and Bias Correction Method (BiC)~\cite{Wu_2019_CVPR}. We also compare against Fixed representations (\emph{FixedRep}) and Fine tuning (\emph{FineTune}). \emph{iTAML}, achieves state-of-the-art incremental learning performance on the wide range of datasets. For MNIST and SVHN, results in Table~\ref{tbl:mnsit_svhn} show that \emph{iTAML} achieves classification accuracy of $97.95\%$ and $93.97\%$ respectively. This is an absolute improvement of $1.79\%, 5.06\%$ over the second best method. 

\begin{table}[t!]
\small
    \centering
            \begin{center}
            \begin{tabular}{l c c}
            \toprule
          \textbf{Methods} & \textbf{MNIST}($A_5$) & \textbf{SVHN}($A_5$)  \\
             \midrule
              EWC \cite{kirkpatrick2017overcoming} & 19.80\%  &  18.21\% \\
            Online-EWC  \cite{schwarz2018progress}  & 19.77\%  & 18.50\%  \\
            SI   \cite{zenke2017continual}     & 19.67\%  &  17.33\% \\
            MAS  \cite{aljundi2018memory}     & 19.52\%  &  17.32\% \\
            LwF  \cite{li2018learning}     & 24.17\%  &  - \\
            \midrule
            GEM$^*$ \cite{lopez2017gradient}       &   92.20\%  &  75.61\% \\
             DGR$^*$ \cite{shin2017continual}       &   91.24\%  & -  \\
            RtF$^*$ \cite{van2018generative}       &   92.56\%  & - \\
            RPS-net$^{*}$\cite{rps_net}  &   {96.16\%}  & {88.91\%}  \\
            \midrule
            Ours$^{*}$  &   \textbf{{97.95\%}}  & \textbf{{93.97\%}}  \\  
            \bottomrule
            \end{tabular}
            \end{center}\vspace{-1.5em}
            \caption{\emph{Comparison on MNIST and SVHN datasets.}  `$*$' denotes memory based methods. \emph{iTAML} outperforms state-of-the-art and performs quite close to oracle case.}
            \label{tbl:mnsit_svhn}
\end{table}

\begin{figure*}[h!]
    \centering
    \begin{subfigure}{0.32\textwidth}
        \centering
        \includegraphics[width=1\textwidth]{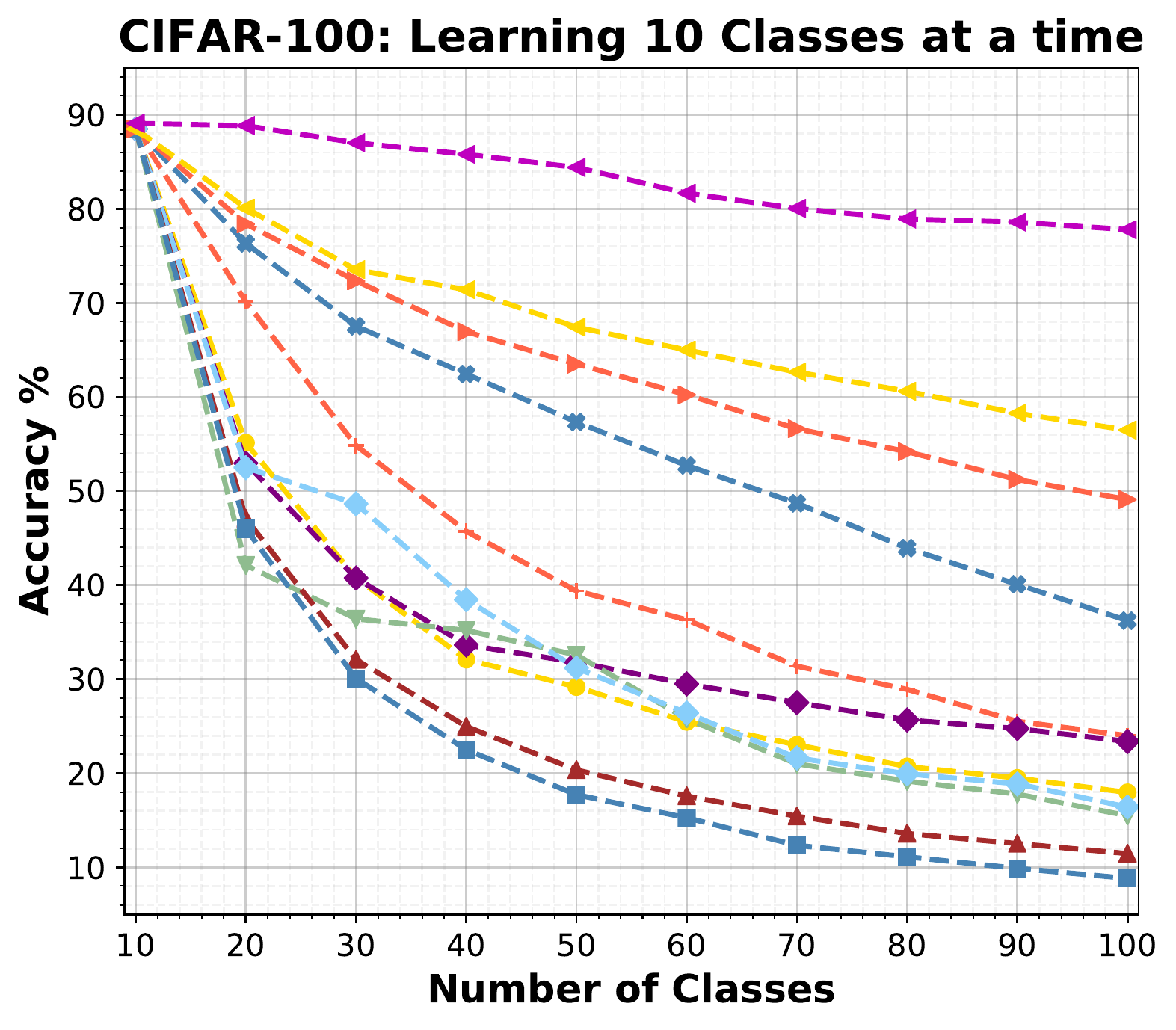}
    \end{subfigure}%
    \begin{subfigure}{0.32\textwidth}
        \centering
        \includegraphics[width=1\textwidth]{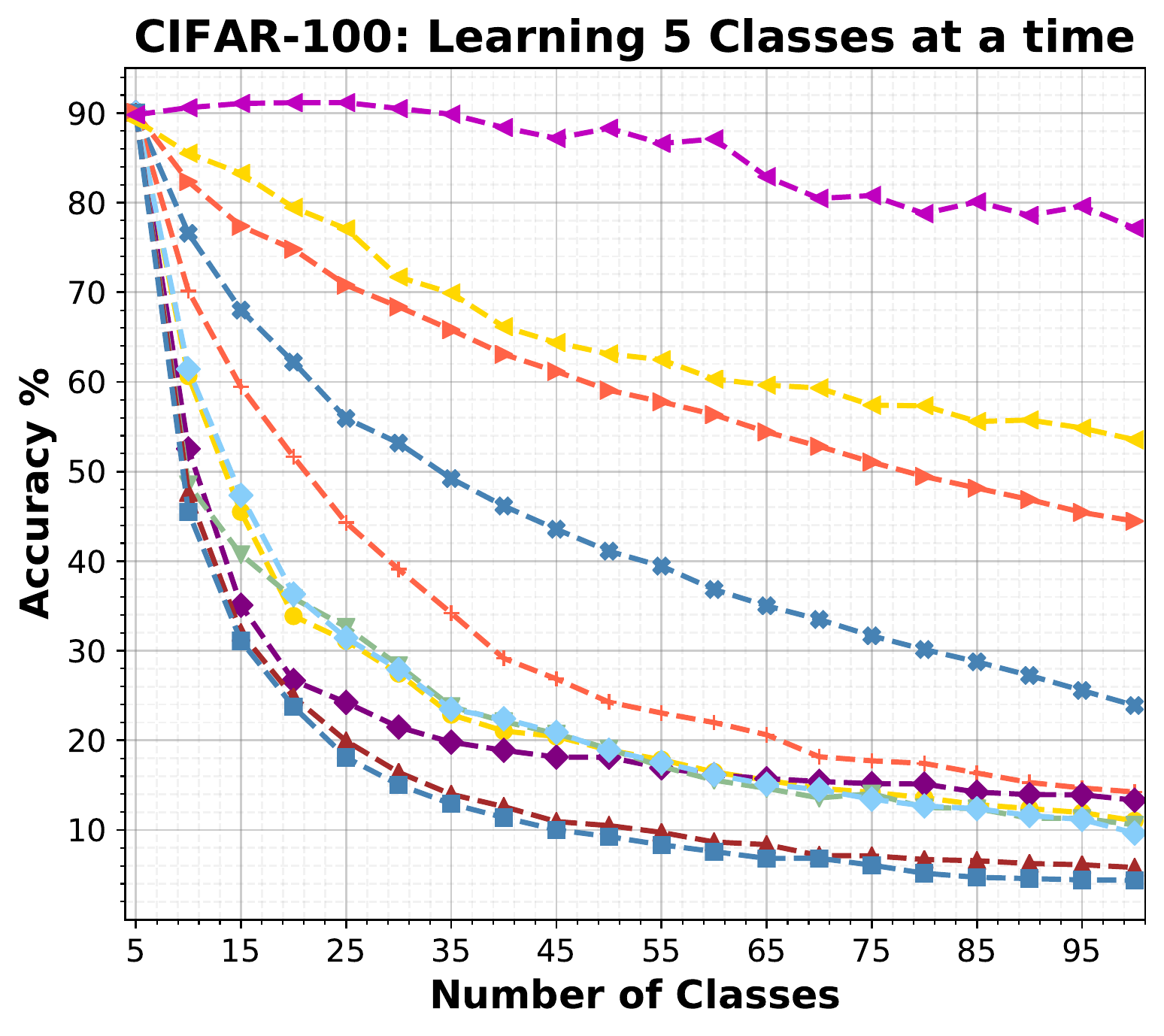}
    \end{subfigure}%
    \begin{subfigure}{0.32\textwidth}
        \centering
        \includegraphics[width=1\textwidth]{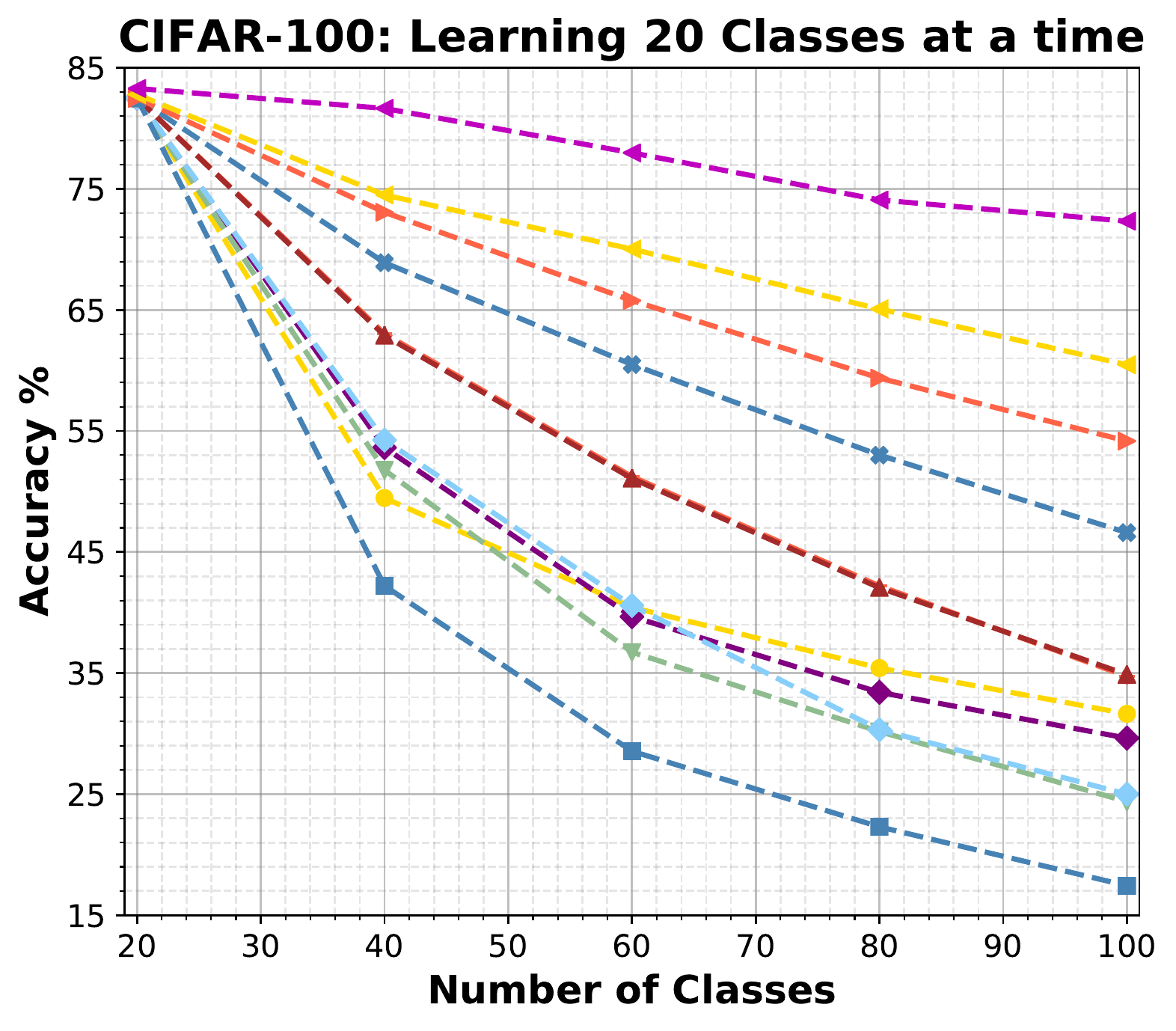}
    \end{subfigure} 
     \\
    \begin{subfigure}{1\textwidth}
        \centering
        \includegraphics[scale=0.5, angle=-90]{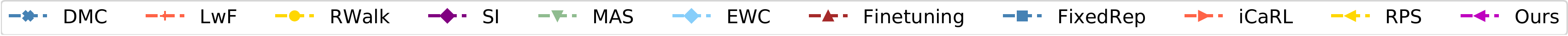}
        \vspace{-0.2cm}
    \end{subfigure}%
    \caption{Classification accuracy on CIFAR100, with $10,20$ and $5$ tasks from left to right. \emph{iTAML} consistently outperforms the existing state-of-the-art across all settings.}
    \label{fig:cifar}
\end{figure*}

Fig.~\ref{fig:cifar} compares different methods on CIFAR100, for incrementally adding tasks with  $10,5$ and $20$ classes at once, by keeping $p=20$. \emph{iTAML} consistently achieves state-of-the-art results across all settings and outperforms the existing methods by a large margin. For incrementally learning $10$ tasks, \emph{iTAML} surpasses the current state-of-the-art RPS-net~\cite{rps_net} by a margin of $21.3\%$. Similarly, we achieve gains of $23.6$ and $18.2\%$ on incrementally learning $5$ and $20$ classes respectively. Results on large scale datasets are shown in Table~\ref{tbl:imagenet}. For ImageNet-100, ImageNet-1K and MS-Celeb-10K datasets, we keep data continuum size as $50,100$ and $20$ respectively. We achieve $89.8\%$ on ImageNet-100 and surpass the current best method by a margin of $15.7\%$. Similarly, on ImageNet-1K, \emph{iTAML} achieves $63.2\%$ with an absolute gain of $19.1\%$. On MS-Celeb-10K dataset with $10,000$ classes, the proposed \emph{iTAML} achieves $95.02\%$ accuracy and retains its performance with addition of new classes. These experiments strongly demonstrate the suitability and effectiveness of \emph{iTAML} for large scale incremental learning tasks.  
\setlength{\tabcolsep}{0.2cm}
\begin{table*}[h]
\footnotesize
\setlength\extrarowheight{-3pt}
    \centering
            \begin{center}
            \begin{tabular}{c l c c c c c c c c c c}
\toprule[0.4mm]
\textbf{Datasets} & \textbf{Methods} & \textbf{1} & \textbf{2}& \textbf{3}& \textbf{4}& \textbf{5}& \textbf{6}& \textbf{7}& \textbf{8}& \textbf{9}& \textbf{Final} \\
             \midrule
\multirow{6}{*}{ImageNet-100/10}
& Finetuning & 99.3&49.4&32.6&24.7&20.0&16.7&13.9&12.3&11.1&9.9 \\
& FixedRep & 99.3&88.1&73.7&62.6&55.7&50.2&42.9&41.3&39.2&35.3 \\
& LwF(TPAMI'18)\cite{li2018learning} & 99.3&95.2&85.9&73.9&63.7&54.8&50.1&44.5&40.7&36.7 \\
& iCaRL(CVPR'17)\cite{rebuffi2017icarl} & 99.3&97.2&93.5&91.0&87.5&82.1&77.1&72.8&67.1&63.5 \\
& RPSnet(NeurIPS'19)\cite{rps_net} & 100.0&97.4&94.3&92.7&89.4&86.6&83.9&82.4&79.4&74.1 \\
& Ours & 99.4&96.4&94.4&93.0&92.4&90.6&89.9&90.3&90.3& \ \ \ \ \ \ \ \ \ \  $\textbf{89.8}_{\textcolor{red}{+15.7}}$ \\
\midrule
\multirow{5}{*}{ImageNet-1K/10}
& Finetuning & 90.2&43.1&27.9&18.9&15.6&14.0&11.7&10.0&8.9&8.2 \\
& FixedRep & 90.1&76.1&66.9&58.8&52.9&48.9&46.1&43.1&41.2&38.5 \\
& LwF(TPAMI'18)\cite{li2018learning} & 90.2&77.6&63.6&51.6&42.8&35.5&31.5&28.4&26.1&24.2 \\
& iCaRL(CVPR'17)\cite{rebuffi2017icarl} & 90.1&82.8&76.1&69.8&63.3&57.2&53.5&49.8&46.7&44.1 \\
& Ours & 91.5&89.0&85.7&84.0&80.1&76.7&70.2&71.0&67.9& \ \ \ \ \ \ \ \ \ \  $\textbf{63.2}_{\textcolor{red}{+19.1}}$ \\
\midrule
\multirow{4}{*}{MS-Celeb-10K/10}
& iCaRL(CVPR'17)\cite{rebuffi2017icarl} & 94.2&93.7&90.8&86.5&80.8&77.2&74.9&71.1&68.5&65.5 \\
& RPSnet(NeurIPS'19)\cite{rps_net} & 92.8&92.0&92.3&90.8&86.3&83.6&80.0&76.4&71.8&65.0 \\
& BiC(CVPR'19)\cite{Wu_2019_CVPR} & 95.7&96.5&96.5&95.7&95.1&94.2&93.2&91.7&90.0&87.6 \\
& Ours & 94.0&95.6&96.0&95.8&95.5&95.4&95.2&95.1&95.0&  \ \ \ \ \ \ \ \ $\textbf{95.0}_{\textcolor{red}{+7.4}}$ \\
\bottomrule[0.4mm]
            \end{tabular}
            \end{center}\vspace{-2em}
            \setlength{\belowcaptionskip}{-6pt}
            \caption{Large-scale experiments on ImageNet-1K and and MS-Celeb-10K show that \emph{iTAML} outperforms all the state-of-the-art methods by a {significant} margin. Note that reported task $t$ accuracy is an average of all $1,2, .., t$ tasks.}
            \label{tbl:imagenet}
\end{table*}
\begin{figure*}[h!]
    \centering
    \begin{subfigure}{0.245\textwidth}
        \centering
        \includegraphics[width=0.99\textwidth]{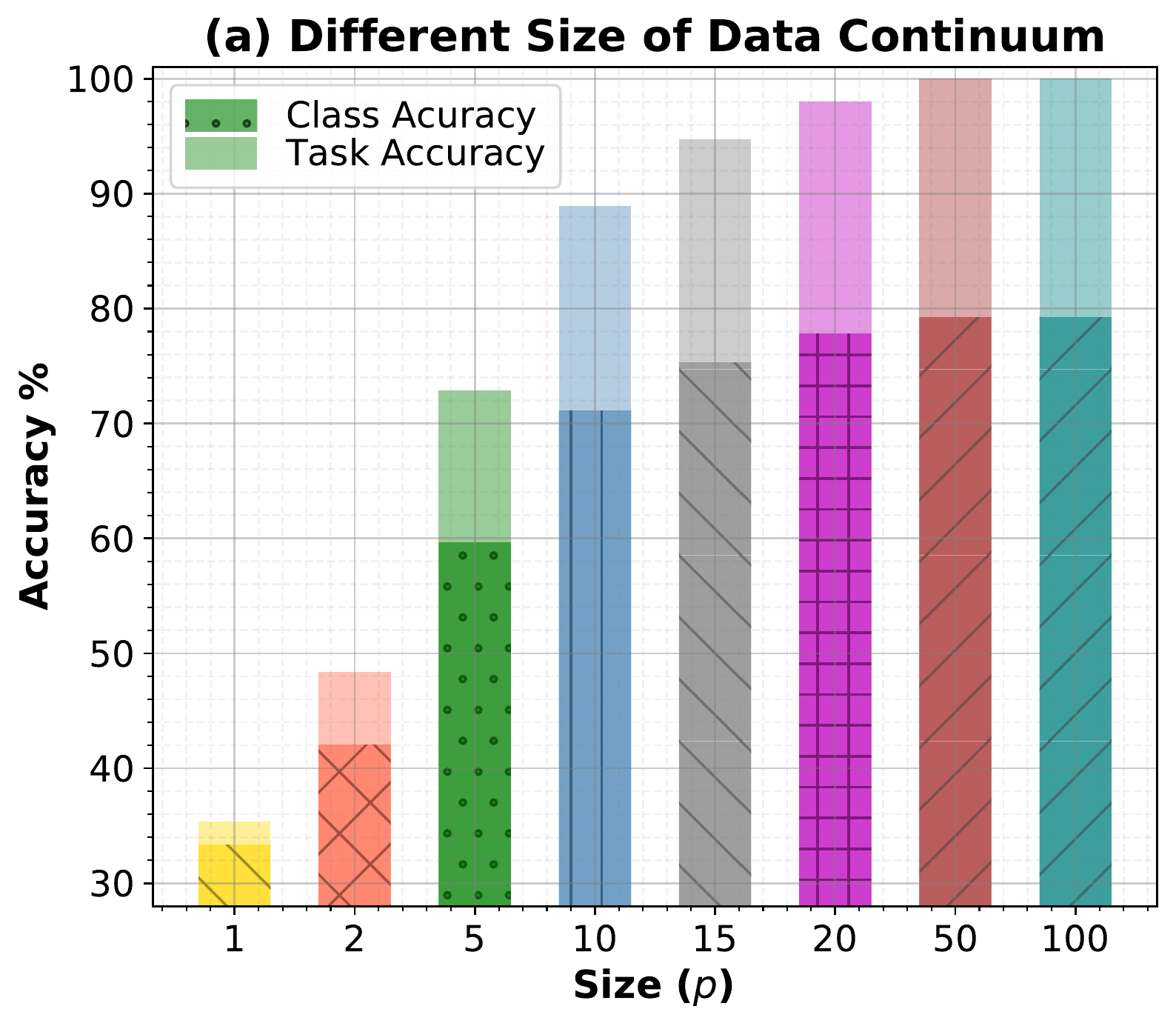}
        \label{fig:ab_continuum_size}
        \vspace{-0.5cm}
    \end{subfigure}%
    \begin{subfigure}{0.245\textwidth}
        \centering
        \includegraphics[width=0.99\textwidth]{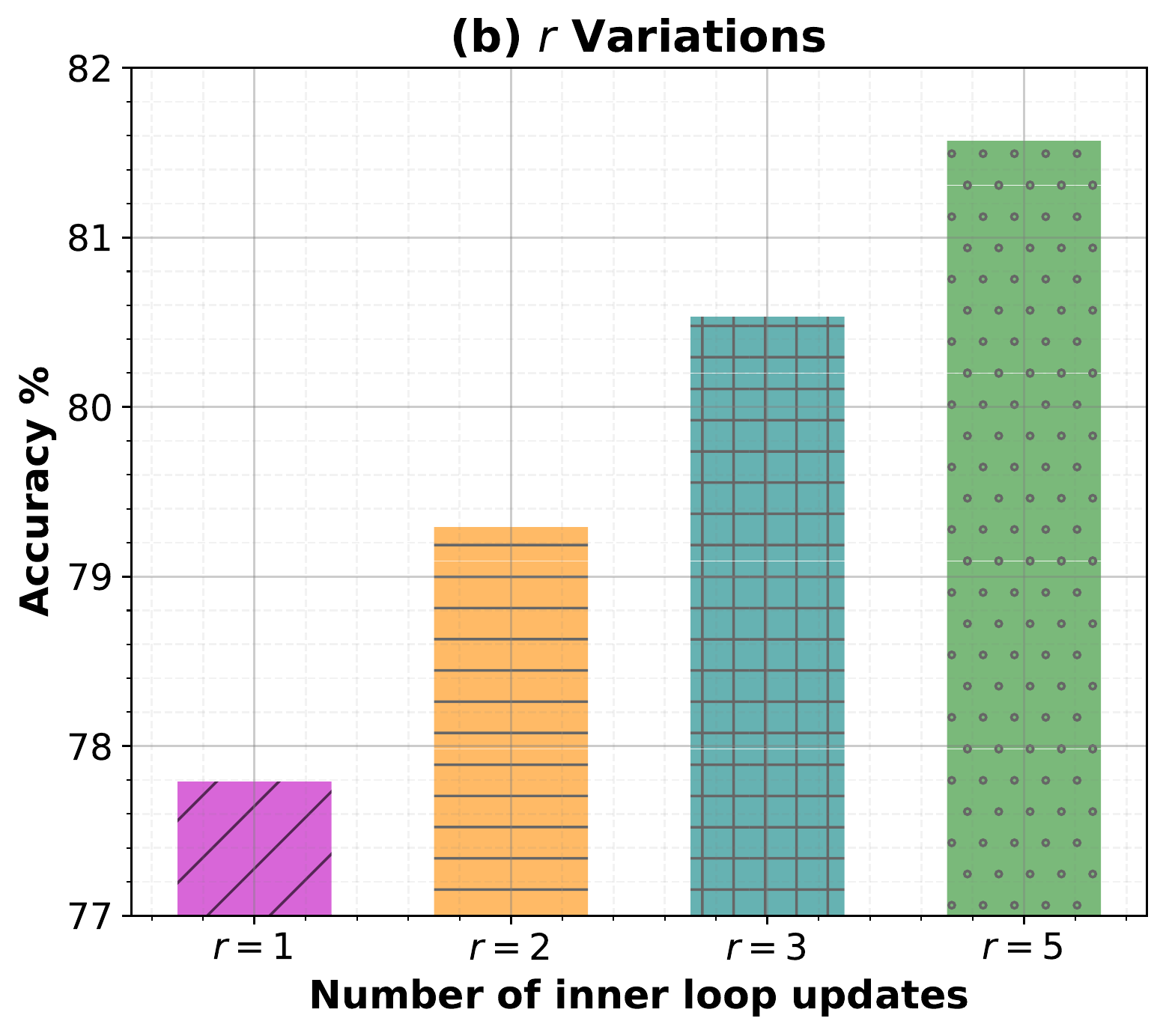}
        \label{fig:ab_inner_updates}
        \vspace{-0.5cm}
    \end{subfigure}%
    \begin{subfigure}{0.245\textwidth}
        \centering
        \includegraphics[width=0.99\textwidth]{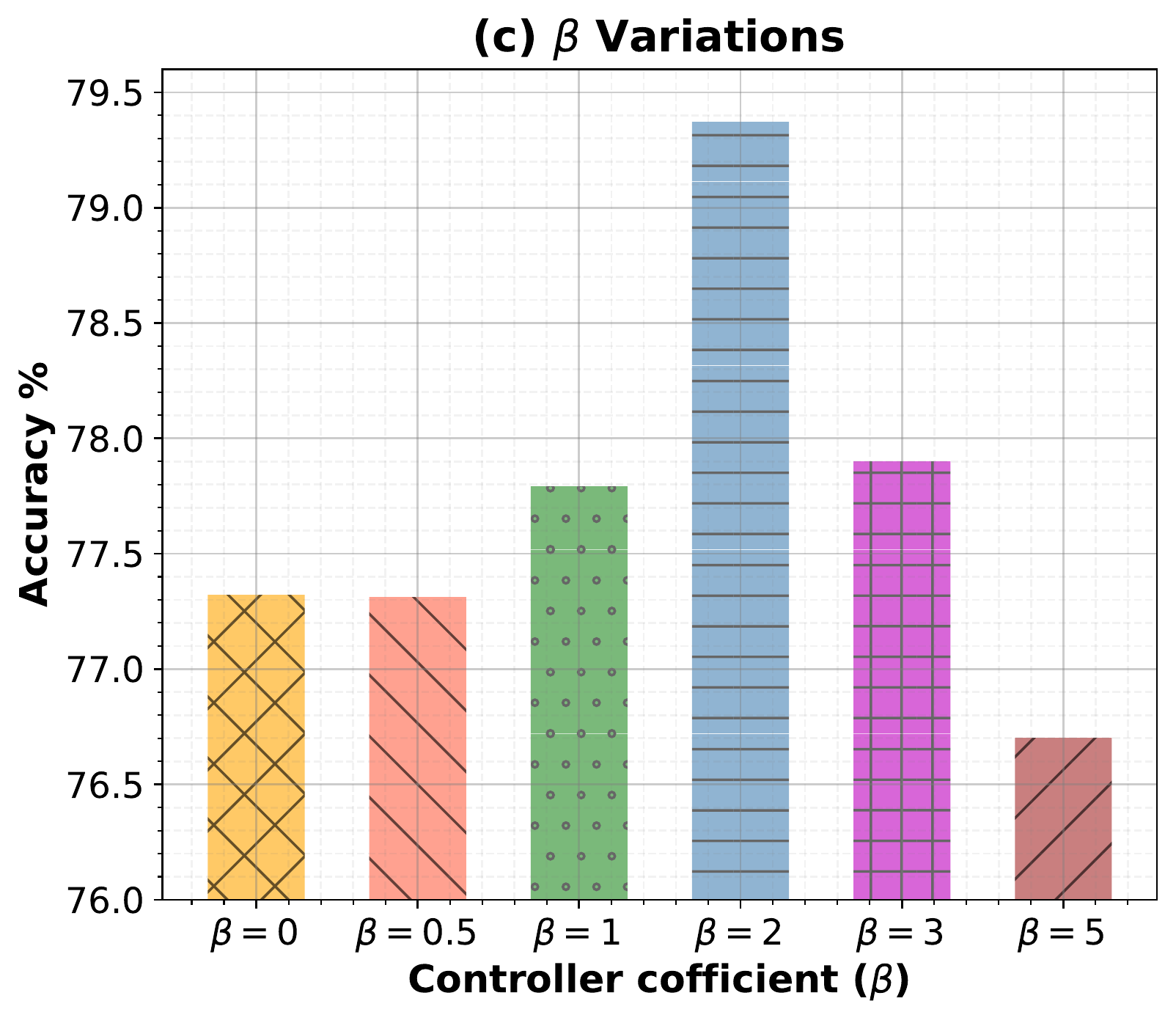}
        \label{fig:ab_beta}
        \vspace{-0.5cm}
    \end{subfigure} 
    \begin{subfigure}{0.245\textwidth}
        \vspace{0.2cm}
        \centering
        \includegraphics[width=0.99\textwidth]{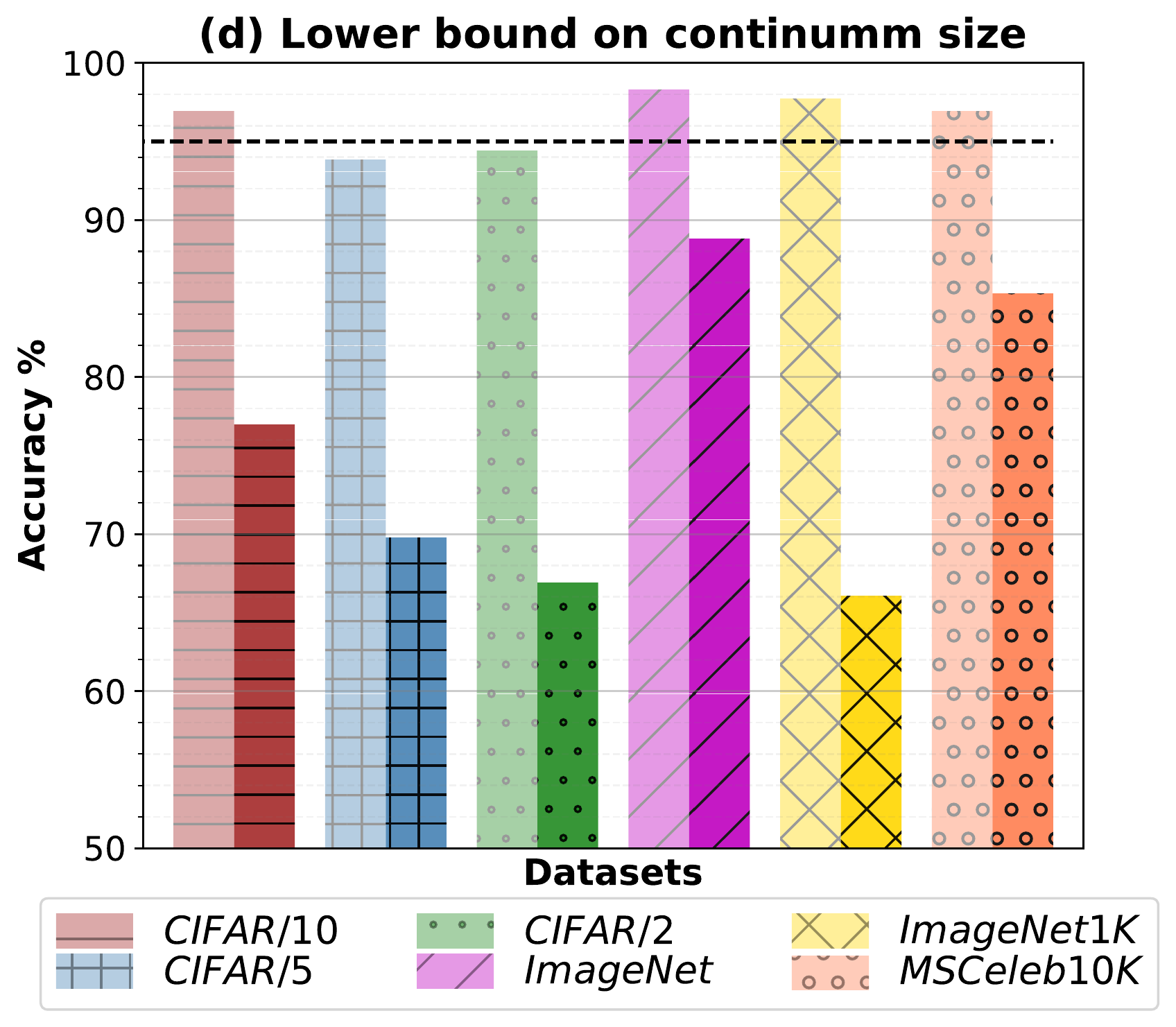}
        \label{fig:ab_backends}
        \vspace{-0.5cm}
    \end{subfigure} 
    \vspace{-0.1cm}
    \setlength{\belowcaptionskip}{-12pt}
    \caption{\emph{Ablation studies:} We study impact of different settings on \emph{iTAML}'s performance: \textbf{a)} Size of the data continuum at inference time vs task and class accuracies, \textbf{b)} Number of inner loop updates $r$, \textbf{c)} Variation in hyper parameters $\beta$, and \textbf{d)} Task and class accuracy of the model on various datasets with a data continuum size taken from Algorithm~\ref{alg:sample} with $\gamma=95\%$.}
    \label{fig:ablations}
\end{figure*}
\vspace{-0.1cm}
\noindent\subsection{Ablation Analysis}
\vspace{-0.1cm}
We perform extensive ablation studies for different aspects of \emph{iTAML} using CIFAR100 with 10 tasks. 

\noindent\textbf{Data Continuum Size ($p$):} At inference, given a data continuum, \emph{iTAML} predicts the task, and individual sample class labels. Here, we study the impact of number of samples in the continuum on task prediction accuracy. Fig.~\ref{fig:ablations}(a) shows that performance improves with $p$. This is because, with a higher number of samples, the noise in the average response is attenuated, thereby improving task accuracy. However, the gain in the task prediction accuracy increases logarithmically. Thus, sufficient value for $p$ ranges from $20$ for CIFAR100 to as low as $3$ for MS-Celeb-10K. 

\noindent\textbf{Variations in $r$:} For higher number of inner gradient updates $r$, the model in the inner loop goes close to the task-specific optimal solution manifold, and the meta-model becomes more ``diverse". We can see this pattern in Fig.~\ref{fig:ablations}(b). With $r=5$, \emph{iTAML} achieves $81.57\%$ while with $r=1$, it achieves $77.79\%$. However, for a model which has seen $T$ tasks, the number of gradient updates in a batch will be $\mathcal{O}(T\cdot r)$. This slows down the training with new incoming tasks. Therefore, we keep $r=1$, as a good trade-off between performance and computational complexity.

\noindent\textbf{Variations in $\beta$:} The parameter $\beta$ controls the speed of learning new information i.e., for higher $\beta$ the model does not learn any new information, and with smaller $\beta$ it only learns the new information and forgets the old one. Fig.~\ref{fig:ablations}(c) shows that the performance improves as we vary $\beta=0$ to $\beta=2$, since this enhances model's ability to remember old knowledge. However, for larger $\beta$, the performance drops as model's stability increase and it is unable to learn new knowledge. We keep $\beta=1$ in our experiments.

\noindent\textbf{Lower bound on Data Continuum:} We set $\gamma=95\%$ and find the required value for $n$ in Algorithm~\ref{alg:sample}. This value is used as the data continuum size during inference for task prediction. As shown in Fig.~\ref{fig:ablations}(d), all the datasets achieve task accuracy around $95\%$, varying from $n=17$ for CIFAR100 to $n=3$ for MS-Celeb.

\vspace{-0.1cm}
\section{Conclusion}
\vspace{-0.1cm}
Incremental learning aims to learn a single model that can continuously adapt itself as the new information becomes available, without overriding existing knowledge. To this end, this work proposes to update the model such that a common set of parameters is optimized on all so-far-seen tasks, without being specific to a single learning task.  We develop a meta-learning approach to train a generic model that can be fast updated for a specific task. In our design, we ensure a balanced update strategy that keeps an equilibrium between old and new task information. Our approach is task-agnostic, and can automatically detect the task at hand, consequently updating itself to perform well on the given inputs. Our experiments demonstrate consistent improvements across a range of classification datasets including ImageNet, CIFAR100, MNIST, SVHN and MS-Celeb.

{\small
\bibliographystyle{ieee_fullname}
\bibliography{ref}
}
\clearpage
\setcounter{section}{0}
\def\thesection{Appendix \Alph{section}}

\section{Supplementary Materials}

\subsection{iTAML vs Other Meta Algorithms}

\begin{customthm}{1}
Given a set of feature space parameters $\theta$ and task classification parameters $\phi = \{\phi_1, \phi_2, \dots \phi_T\}$, after $r$ inner loop updates, iTAML's meta update gradient for task $i$ is given by,
\begin{align*}
    \mathlarger{g}_{itaml}(i) = \gradx{0}{i,0} + \dots + \gradx{0}{i,r-1}, 
\end{align*}
where, $\gradx{0}{i,j}$ is the $j^{th}$ gradient update with respect to $\{\theta,\phi_i\}$ on a  single micro-batch.
\end{customthm}
\begin{proof}
\label{proof:1}
  Let $\Phi_i = \{\theta,\phi_i\}$ is the set of feature-space parameters and task-specific parameters of the task $i$, $\mathcal{L}_i(\Phi_i)$ is the loss calculated on a specific micro-batch $\mathcal{B}^i_{\mu}$ for task $i$ using $\Phi_i$, and $\alpha$ is the inner loop learning rate. The parameters update is given by,
  \begin{align*}
    \Phi_{i,r} = \Phi_{i,r-1} - \alpha \nabla_{\Phi_{i,r-1}} \mathcal{L}_i(\Phi_{i,r-1}), \text{ where } \Phi_{i,0} = \Phi_i .
\end{align*}
Lets take $\gradx{0}{i,j} =  \nabla_{\Phi_{i,j}}\mathcal{L}_i(\Phi_{i,j})$, 
\begin{align*}
    \Phi_{i,r} &= \Phi_{i,r-1} - \alpha \gradx{0}{i,r-1}.
\end{align*}
Using the meta gradient update rule defined in  Reptile~\cite{nichol2018reptile} i.e., $ (\theta_{i,0} - \theta_{i,r})/\alpha$, we have,
\begin{align*}
    \mathlarger{g}_{itaml}(i) &= \frac{\theta_{i,0} - \theta_{i,r}}{\alpha}\\
                     &= \frac{\theta_{i,0} - (\theta_{i,r-1} - \alpha\gradx{0}{i,r-1})}{\alpha} \\
                    &\;\;\vdots \notag \\  
                    &= \frac{\theta_{i,0} - (\theta_{i,0} - \alpha\gradx{0}{i,0} - \dots - \alpha\gradx{0}{i,r-1})}{\alpha} \\
                    &= \gradx{0}{i,0} + \gradx{0}{i,1} + \dots + \gradx{0}{i,r-1}
\end{align*}
\end{proof}

\begin{customthm}{2}
Given a set of feature space parameters $\theta$ and task classification parameters $\phi = \{\phi_1, \phi_2, \dots \phi_T\}$, iTAML allows to keep the number of inner loop updates $r\geq1$. 
\end{customthm}
\begin{proof}
\label{proof:2}
For a given task $t$, there will be $t$ gradients available for meta update,  
\begin{align*}
    \mathlarger{g}_{itaml} &= \eta \frac{1}{t} \sum_{i=1}^{t} \mathlarger{g}_{itaml}(i) \\
    &=  \exp \left(-\beta\frac{t}{T}\right) \cdot \frac{1}{t} \cdot \sum_{i=1}^{t} \sum_{j=1}^{r-1} \gradx{0}{i,j}.
\end{align*}
Reptile algorithm requires $r>1$ since, $r=1$ would result in joint training in Reptile algorithm. Reptile updates the parameters with respect to $\{\theta, \phi\}$ in the inner loop, while iTAML updates the parameters with respect to $\{\theta, \phi_i\}$ in the inner loop of task $i$. When $r=1$,
\begin{align*}
    \mathlarger{g}_{itaml} &= \exp \left(-\beta\frac{t}{T}\right) \cdot \frac{1}{t} \cdot  \sum_{i=1}^{t} \gradx{0}{i,0}   \\
                        &= \exp \left(-\beta\frac{t}{T}\right) \cdot \frac{1}{t} \cdot  \sum_{i}^{t} \nabla_{\Phi_{i,0}}\mathcal{L}_i(\Phi_{i,0}) \\
                        &= \underbrace{\exp \left(-\beta\frac{t}{T}\right)}_{\text{decaying factor}} \cdot \frac{1}{t} \cdot  \sum_{i=1}^{t} \underbrace{\nabla_{\{\theta,\phi_i\}}\mathcal{L}_i(\{\theta,\phi_i\})}_{\text{task-specific gradient}} \\
                        & \neq \frac{1}{t} \sum_{i=1}^{t} \nabla_{\{\theta,\phi\}}\mathcal{L}_i(\{\theta,\phi\}) = \mathlarger{g}_{joint}
\end{align*}
\end{proof}



\subsection{Additional Results}


\begin{figure}[h]
    \centering
    \begin{subfigure}{0.45\textwidth}
        \centering
        \includegraphics[width=0.99\textwidth]{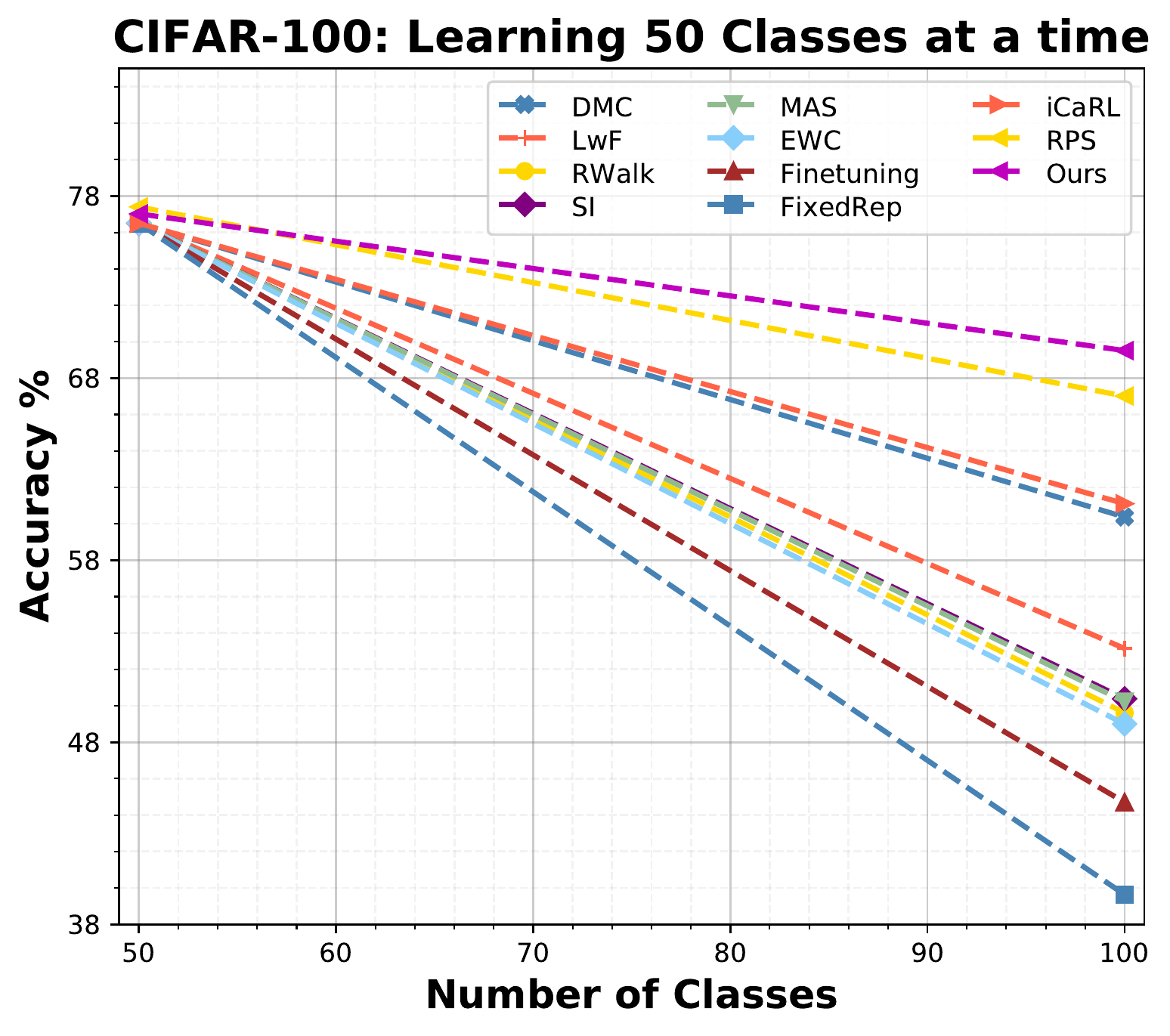}
        \vspace{-0.5cm}
    \end{subfigure} 
    \vspace{-0.2cm}
    \caption{Classification accuracy on CIFAR100, with $2$ tasks. Exemplar memory is set to 2000 samples and \emph{ResNet-18(1/3)} is used for training. We keep $p=20$ for experiments on data continuum.}
    \label{fig:cifar100_2}
\end{figure}

\noindent\textbf{Variation on $b$:} iTAML uses a low $b$ value i.e., $b$=1. Parameter $b$ denotes the number of epochs for model update during adaptation. We observed that higher $b$ values do not have a significant impact on performance, but the time complexity increases linearly with $b$. Below, we report experimental results by changing  $b$ from 1 to 5 and note that the accuracies does not improve significantly. 
\begin{table}[h]
\footnotesize
    \centering
    \setlength\extrarowheight{-3pt}
    \begin{tabular}{c c c c c c}
  \toprule
      $b$      &    1       &  2    & 3    &   4  & 5   \\ \midrule
      Accuracy &  78.24\%   &   78.48\%    &   78.48\%   &   78.53\%   &   78.50\%  \\ \bottomrule
    \end{tabular}
\end{table}

\noindent\textbf{Note on SVHN:} For SVHN dataset, we keep $r=4$ for the last task. This is due to the fact that, SVHN has a lower variance in the data distribution and which forces the model to stuck at the early stages of local minima.

\noindent\textbf{Backends and Optimizers:} We evaluate our method with various architectural backends. Even with a very small model having ($0.49M$) parameters, iTAML can achieve $69.94\%$ accuracy, with a gain of $13.46\%$ over second-best (RPS-net $77.5M$) method. ResNet-18 full model gives $80.27\%$. Further, iTAML is a modular algorithm, we can plug any optimizer into it. We evaluate iTAML with SGD, Adam~\cite{kingma2014adam} and RAdam~\cite{liu2019variance}, and respectively achieve a classification accuracy of $70.34\%, 74.83\%$ and $76.63\%$ with these optimizers.

\end{document}